\documentclass[journal]{IEEEtran}
\usepackage{graphicx}
\usepackage{cite}
\usepackage{amssymb}
\usepackage{amsmath}
\usepackage{enumerate}
\usepackage{booktabs}
\usepackage{algorithm}
\usepackage{algorithmic}
\usepackage{color}
\usepackage{colortbl}
\usepackage[colorlinks,linkcolor=red,anchorcolor=blue,citecolor=green]{hyperref}
\usepackage{bbm}

\usepackage{multirow}
\usepackage[table,xcdraw]{xcolor}

\begin{document}
%
\title{Universal Adversarial Examples in Remote Sensing: Methodology and Benchmark}
%
%
%

\author{Yonghao~Xu,~\IEEEmembership{Member,~IEEE,}
        and~Pedram~Ghamisi,~\IEEEmembership{Senior Member,~IEEE}
\thanks{Y. Xu is with the Institute of Advanced Research in Artificial Intelligence (IARAI), 1030 Vienna, Austria (e-mail: yonghao.xu@iarai.ac.at).}
\thanks{P. Ghamisi is with the Institute of Advanced Research in Artificial Intelligence (IARAI), 1030 Vienna, Austria, and also with Helmholtz-Zentrum Dresden-Rossendorf, Helmholtz Institute Freiberg for Resource Technology, Machine Learning Group, 09599 Freiberg, Germany (e-mail: pedram.ghamisi@iarai.ac.at; p.ghamisi@hzdr.de).}
}

%
%

\markboth{Journal of \LaTeX\ Class Files,~Vol.~xx, No.~xx, March~2022}%
{Shell \MakeLowercase{et al.}: Bare Demo of IEEEtran.cls for IEEE Journals}
%



\maketitle

\begin{abstract}
Deep neural networks have achieved great success in many important remote sensing tasks. Nevertheless, their vulnerability to adversarial examples should not be neglected. In this study, we systematically analyze the universal adversarial examples in remote sensing data for the first time, without any knowledge from the victim model. Specifically, we propose a novel black-box adversarial attack method, namely Mixup-Attack, and its simple variant Mixcut-Attack, for remote sensing data. The key idea of the proposed methods is to find common vulnerabilities among different networks by attacking the features in the shallow layer of a given surrogate model. Despite their simplicity, the proposed methods can generate transferable adversarial examples that deceive most of the state-of-the-art deep neural networks in both scene classification and semantic segmentation tasks with high success rates. We further provide the generated universal adversarial examples in the dataset named UAE-RS, which is the first dataset that provides black-box adversarial samples in the remote sensing field. We hope UAE-RS may serve as a benchmark that helps researchers to design deep neural networks with strong resistance toward adversarial attacks in the remote sensing field. Codes and the UAE-RS dataset are available online (https://github.com/YonghaoXu/UAE-RS).

\end{abstract}


%
\IEEEpeerreviewmaketitle

\section{Introduction}

\IEEEPARstart{R}{ecent} advances in remote sensing have brought about the explosive growth of Earth observation data collected by numerous satellite or airborne sensors \cite{boukabara2021earth}. The massive availability of these data has significantly boosted the development of many important applications in the geoscience and remote sensing field \cite{ghamisi2019multisource,dfc}. Some representative applications include scene classification \cite{cheng2017remote}, object detection \cite{wu2021deep}, and semantic segmentation \cite{ghorbanzadeh2021transferable}. Currently, most of the state-of-the-art methods for these tasks are based on deep neural networks. On the one hand, deep neural networks--especially convolutional neural networks (CNNs)--have achieved great success in the interpretation of remote sensing data \cite{zhang2016deep}. On the other hand, their vulnerability toward adversarial examples should not be neglected.

\begin{figure}
  \centering
  \includegraphics[width=\linewidth]{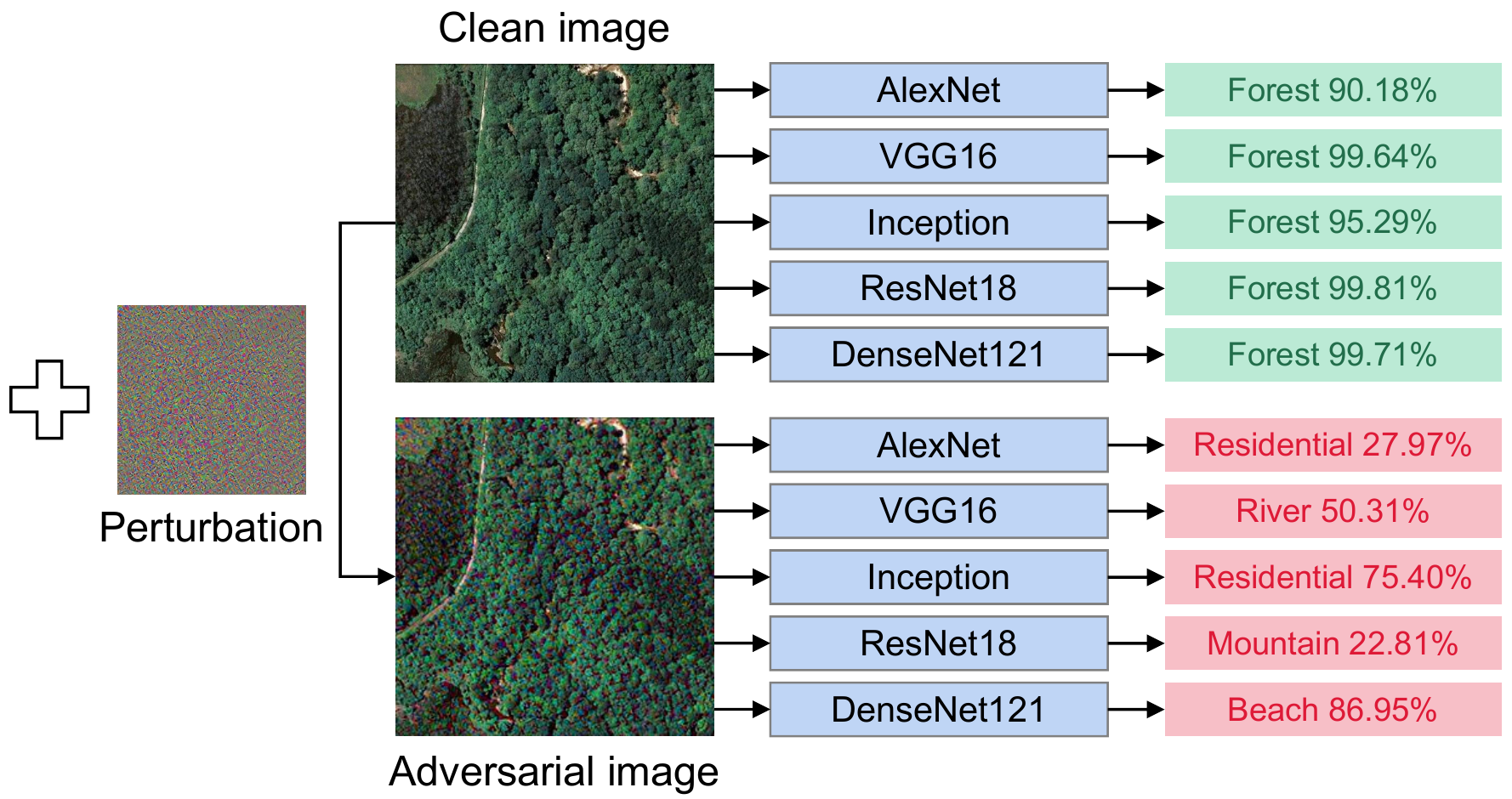}
  \caption{Illustration of the black-box adversarial attack on the scene classification of very high-resolution remote sensing images using the proposed Mixup-Attack (with ResNet18 as the surrogate model). Without accessing any information from the target model, the generated adversarial image can fool different deep neural networks to make wrong predictions.}
\label{fig:intro}
\end{figure}

In \cite{szegedy2013}, Szegedy et al. first discovered that deep neural networks are very fragile to adversarial examples, which can be simply generated by adding subtle adversarial perturbations to the original image. Such perturbations can be produced via specific adversarial attack methods like the box-constrained L-BFGS \cite{szegedy2013} and fast gradient sign method (FGSM) \cite{goodfellow2014}. With these well-designed algorithms, the generated adversarial examples may look very similar to the original images to the human visual system but can mislead state-of-the-art deep neural networks to make wrong predictions with high confidence \cite{akhtar2021advances}. This phenomenon will undoubtedly result in a threat to the security of deep learning-based image recognition systems in real-world applications \cite{kurakin2016adversarial,madry2017towards}. For example, Komkov et al. successfully attacked the advanced face identification system using a rectangular paper sticker generated by an algorithm for off-plane transformations \cite{komkov2019advhat}. Thus, to improve the reliability of deep learning models, it is necessary to study the characteristics of adversarial examples in depth.

In addition to the aforementioned research that focuses on the computer vision field, there is currently some exploration into adversarial examples in the geoscience and remote sensing field \cite{adv_rs,chen2019adversarial,chan2021demotivate}. Czaja et al. first revealed that adversarial examples also exist in the satellite remote sensing image classification task \cite{czaja2018adversarial}. Their experiments indicated that adversarial attacks on a small patch inside the remote sensing image could fool deep learning models into making wrong predictions. Chen et al. conducted an empirical study of adversarial examples on scene classification of remote sensing images, where both optical and synthetic-aperture radar (SAR) images are analyzed \cite{chen2021empirical}.
Xu et al. further discovered that adversarial examples also exist in the hyperspectral domain \cite{xu2021self}. Their experiments revealed that adversarial attacks can successfully change the spectral reflectance characteristics of adversarial hyperspectral samples.

So far, existing research on adversarial attacks in the geoscience and remote sensing field mainly focuses on white-box attacks \cite{czaja2018adversarial,adv_rs,chen2019adversarial,chen2021empirical}, where it is assumed that the attacker has
complete knowledge of the victim model, including its architecture and parameter values, so as to generate the corresponding adversarial examples. However, in real-world scenarios, it is usually impossible to obtain detailed information about the deployed network \cite{narodytska2017simple}. Under such a circumstance, it is more feasible to conduct a black-box attack, where the adversarial examples are generated without any knowledge of the victim model \cite{akhtar2021advances}. Obviously, it is more challenging to implement black-box attacks \cite{guo2019simple}. One possible solution to achieve this goal is to conduct a white-box adversarial attack with a surrogate model whose complete knowledge is obtainable, and then feed the generated adversarial examples to the unknown target model. Nevertheless, traditional adversarial attack methods like the FGSM are generally designed to fool specific deep neural networks. Thus, the transferability of adversarial examples produced from a particular model to other networks may be limited, especially for those that have architectures that are distinctly different from the model used in the attack. For example, the experimental results in \cite{adv_rs} demonstrated that while adversarial examples generated by the AlexNet can significantly cheat AlexNet itself, other deep neural networks like the ResNet or DenseNet possess strong resistance toward this adversarial attack. Therefore, finding the common vulnerability among different networks and generating transferable adversarial examples are critical challenges.

\begin{figure}
  \centering
  \includegraphics[width=\linewidth]{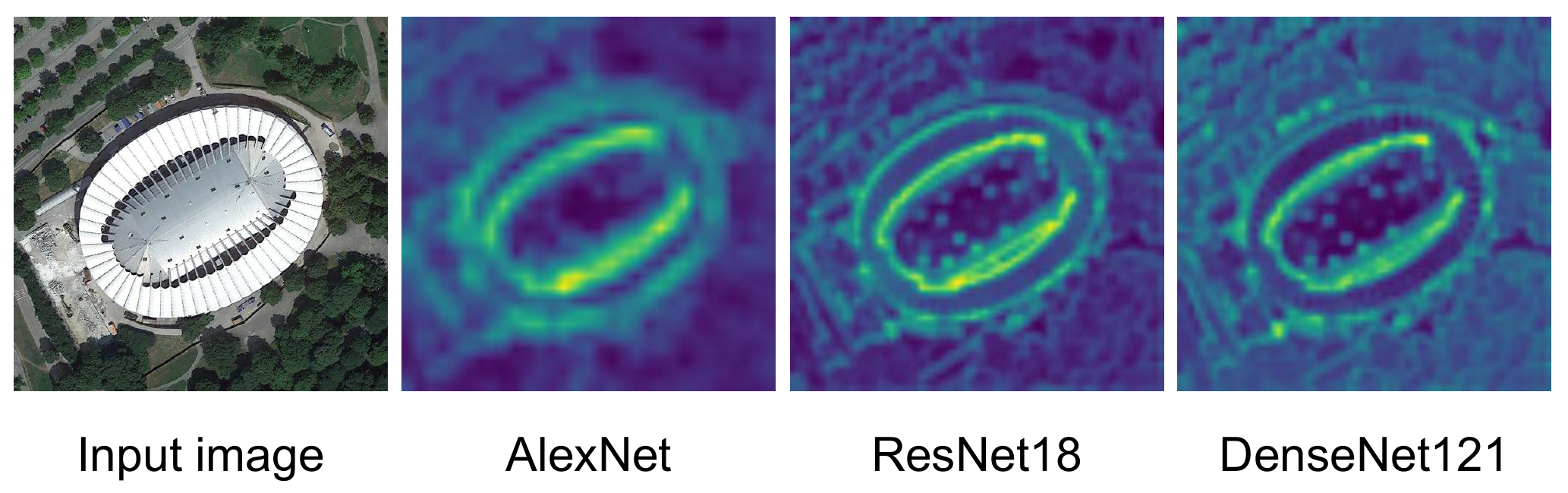}
  \caption{Illustration of the features in the first pooling layer from AlexNet, ResNet18, and DenseNet121. While different networks have distinct architectures, they may yield similar feature representations in the shallow layers.}
\label{fig:feature}
\end{figure}

Based on the above analysis, this study aims to conduct black-box attacks for remote sensing data and generate universal adversarial examples that can achieve a high success rate in attacking different deep neural networks without any knowledge about the victim models, as shown in Fig. \ref{fig:intro}. The initial inspiration of our work comes from an observation that different deep learning models may yield similar feature representations in the shallow layers of the network \cite{yosinski2015understanding}. Compared to the deep features, which are more abstract, features in the shallow layers generally preserve more detailed spatial information in the image and share similar representations, even in different networks (see Fig. \ref{fig:feature} for a visual example). Thus, a natural idea is to implement feature-level adversarial attacks on the shallow layers, which may bring about better transferability to different victim models. To this end, we propose the Mixup-Attack method. Specifically, we first construct the mixup image by the linear combination of images from different categories. Given a surrogate deep neural network, we extract the shallow features of the mixup image and the target image under attack. Then, we define the mix loss function $\mathcal{L}_{mix}$ by minimizing the KL-divergence between features of the mixup image and the input image. Considering the constraint of $\mathcal{L}_{mix}$ does not take the prediction of the network into consideration, the cross-entropy loss $\mathcal{L}_{ce}$ between the predicted logits on the target image and the true label is also adopted to assist the attack. Thus, the complete objective function $\mathcal{L}$ is the weighted combination of $\mathcal{L}_{mix}$ and $\mathcal{L}_{ce}$. Finally, the universal adversarial examples can be produced by adding the gradients (adversarial perturbation) of the complete objective function $\mathcal{L}$ to the original clean image. We further propose a variant of Mixup-Attack called the Mixcut-Attack method, in which the mixup image is simply constructed with slices of images from different categories. Despite their simplicity, we find the proposed methods can generate transferable adversarial examples that cheat most of the state-of-the-art deep neural networks without any knowledge about them.

The main contributions of this paper are summarized as follows.

\begin{enumerate}
\item We systematically analyze the universal adversarial examples in remote sensing data for the first time, without any knowledge about the victim model. Our research reveals the significance of the resistance and robustness of deep learning models when addressing safety-critical remote sensing tasks.
\item We propose a novel black-box adversarial attack method, called Mixup-Attack, and its simple variant Mixcut-Attack, for remote sensing data. Extensive experiments on four benchmark remote sensing datasets verify the effectiveness of the proposed methods in both scene classification and semantic segmentation tasks.
\item We provide the generated universal adversarial examples in a dataset named UAE-RS, which is the first dataset that provides black-box adversarial samples in the remote sensing field. Experiments demonstrate that samples in UAE-RS can mislead the existing state-of-the-art deep neural networks into making wrong predictions with high success rates, which may serve as a benchmark for researchers developing adversarial defenses.
\end{enumerate}

The rest of this paper is organized as follows. Section II reviews works related to this study. Section III describes the proposed adversarial attack methods in detail. Section IV presents the information on datasets used in this study and the experimental results. Conclusions and other discussions are summarized in Section V.

\section{Related Work}
\subsection{Adversarial Attacks}
\subsubsection{Box-Constrained L-BFGS}
The first adversarial attack method was proposed by Szegedy et al., where they generated adversarial perturbations by maximizing the network's prediction error \cite{szegedy2013}.

Formally, let $f: x \in \mathbb{R}^{n}\rightarrow y\in \mathbb{L}$ be the mapping function of a deep neural network that maps an image with $n$ pixels into a discrete label set. Given an image $x$ and a target label $\hat y$, where $\hat y$ denotes the wrong label that we expect the network would predict, the adversarial perturbation $\rho$ can be produced by solving the box-constrained optimization problem as below:
\begin{equation}
    \mathop{\min}_{\rho} \| \rho\|_{2}, subject\  to:
    \begin{cases}
    f\left(x+\rho\right)=\hat y\\
    x+\rho \in \left[0,1\right]^{n}.
    \end{cases}
\label{eq:lbfgs}
\end{equation}
In general, directly solving \eqref{eq:lbfgs} is a hard problem. Szegedy et al. proposed to approximate the solution of \eqref{eq:lbfgs} using a box-constrained L-BFGS. More concretely, they performed line-search to find the minimum $c>0$ for which the minimizer $\rho$ of the following optimization problem satisfies $f\left(x+\rho\right)=\hat y$, and $\hat y\not = y$:
\begin{equation}
\begin{split}
    &\mathop{\min}_{\rho} c\| \rho\|_{2}+J\left(\theta,x+\rho,\hat{y}\right),\\
    &subject\ to: x+\rho \in \left[0,1\right]^{n},
\end{split}
\label{eq:lbfgs2}
\end{equation}
where $\theta$ represents the parameters in the deep neural network, and $J\left(\cdot\right)$ denotes the loss function used for training the network (e.g., the cross-entropy loss).

\subsubsection{Fast Gradient Sign Method}
In practical applications, the optimization of \eqref{eq:lbfgs2} is still very difficult since it requires layer-wise optimization for parameters in different layers. To make adversarial attacks more efficient, Goodfellow et al. proposed the fast gradient sign method (FGSM) \cite{goodfellow2014}. Given an image $x$ and its true label $y$, the adversarial example $x_{adv}$ can be calculated as:
\begin{equation}
x_{adv}={\rm clip}\left(x+\epsilon ~ {\rm sign} \left(\nabla _x J\left(\theta,x,y\right)\right)\right),
\label{eq:fgsm}
\end{equation}
where $\nabla _x J\left(\theta,x,y\right)$ calculates the gradients of the loss function $J\left(\cdot\right)$ with respect to the input sample $x$, ${\rm sign}\left(\cdot\right)$ denotes the sign function, ${\rm clip}\left(\cdot\right)$ clips the pixel values in the image, and $\epsilon$ is a small scalar value that controls the norm of the perturbation.

Miyato et al. \cite{miyato2018virtual} and Kurakin et al. \cite{kurakin2016adversarial} further extend FGSM by applying the $\ell_2$ norm and $\ell_{\infty}$ norm to the generated perturbation:
\begin{equation}
    \ell_{2}:\ x_{adv}={\rm clip}\left(x+\epsilon \frac{\nabla _x J\left(\theta,x,y\right)}{\|\nabla _x J\left(\theta,x,y\right)\|_2}\right).
\label{eq:l2}
\end{equation}
\begin{equation}
    \ell_{\infty}:\ x_{adv}={\rm clip}\left(x+\epsilon \frac{\nabla _x J\left(\theta,x,y\right)}{\|\nabla _x J\left(\theta,x,y\right)\|_{\infty}}\right).
\label{eq:linf}
\end{equation}

\subsubsection{Iterative Fast Gradient Sign Method}
The iterative fast gradient sign method (I-FGSM) was first proposed by Kurakin et al., and is an iterative version of FGSM \cite{kurakin2016adversarial}. At each iteration, the adversarial example can be updated as below:
\begin{equation}
     x_{adv}^{t+1}={\rm clip}\left(x_{adv}^t+\alpha~{\rm sign} \left(\nabla _{x_{adv}} J\left(\theta,x_{adv}^t,y\right)\right) \right),
\label{eq:ifgsm}
\end{equation}
where $\alpha$ is the step size. When $t=0$, $x_{adv}^0$ is initialized with the original clean image $x$.

\subsubsection{Carlini and Wagner's Attack (C\&W)}
Carlini and Wagner proposed to conduct adversarial attacks by encouraging $x_{adv}$ to have a larger probability score for a wrong class than all other classes \cite{carlini2017towards}. This method directly optimizes the distance between the benign examples and the adversarial examples by solving:
\begin{equation}
     \mathop{\arg\min}_{x_{adv}} \|x_{adv}-x\|_\infty-\mu J\left(\theta,x_{adv},y\right),
\label{eq:cw}
\end{equation}
where $\mu$ is a weighting factor. A more comprehensive review of this method can be found in \cite{xu2020adversarial}.

\subsection{Black-Box Attacks}
Since the discovery of adversarial examples, numerous efforts have been made to design advanced adversarial attack methods \cite{akhtar2021advances}. Nevertheless, most of the previous research focuses on the white-box attack, where it is assumed that complete knowledge of the victim model, including its architecture and parameter values, is accessible for the attacker. Although these white-box attack methods can achieve high success rates in the attack, their assumption is usually invalid in practical applications. To make the attack more pragmatic, black-box attack methods are proposed which assume no (or minimal) knowledge about the victim model \cite{akhtar2021advances}. Generally, black-box attacks can be categorized into query-based and transfer-based methods.

\subsubsection{Query-Based Black-Box Attacks}
Query-based black-box attacks assume that the output of the victim network is accessible during the attack \cite{narodytska2017simple,vo2022query}. The main idea is to add subtle perturbations to the input image and observe the output of the victim model. After a series of queries, the real gradients of the victim network can be reconstructed via numerical approximation. For example, Narodytska et al. adopted the greedy local search strategy, where a local neighborhood is used to refine the current image and optimize the objective function related to the network's output in each iteration \cite{zhang2022beyond}. To make the query more efficient, Guo et al. proposed to randomly sample a vector from a predefined orthonormal basis and either add or subtract it to the target image \cite{guo2019simple}. This simple iterative principle can result in higher query efficiency for both targeted and untargeted attacks. Some other representative query strategies include Bayes optimization \cite{ru2019bayesopt}, evolutional algorithms \cite{meunier2019yet}, and meta-learning \cite{du2019query}.

\subsubsection{Transfer-Based Black-Box Attacks}
The main limitation of query-based black-box attacks lies in the fact that a good approximation for the gradients of a victim model is usually very hard considering the complexity of the deployed model in practice \cite{li2020learning}. Thus, it may require a large number of queries to get a high success rate in the attack. By contrast, the main idea of a transfer-based black-box attack is to conduct the white-box attack on a surrogate model and adopt the generated adversarial examples to attack the unknown victim model. Obviously, the success rate of the attack can increase if the surrogate model can share a similar architecture with the victim model. However, since knowledge about the victim model is inaccessible in black-box attacks, directly using traditional attack methods like the FGSM can hardly achieve satisfactory performance. Thus, recent research aims to improve the inherent transferability of the adversarial attack \cite{li2020learning}. For example, Liu et al. found that ensemble learning helps to generate more transferable adversarial examples \cite{liu2016delving}. Zhou et al. and Huang et al. found the attack on the intermediate level is more powerful than the attack on the predicted logits \cite{zhou2018transferable,huang2019enhancing}. Chen et al. proposed to conduct adversarial attacks on the attention maps of input images \cite{chen2020universal}.

\begin{figure*}
  \centering
  \includegraphics[width=0.85\linewidth]{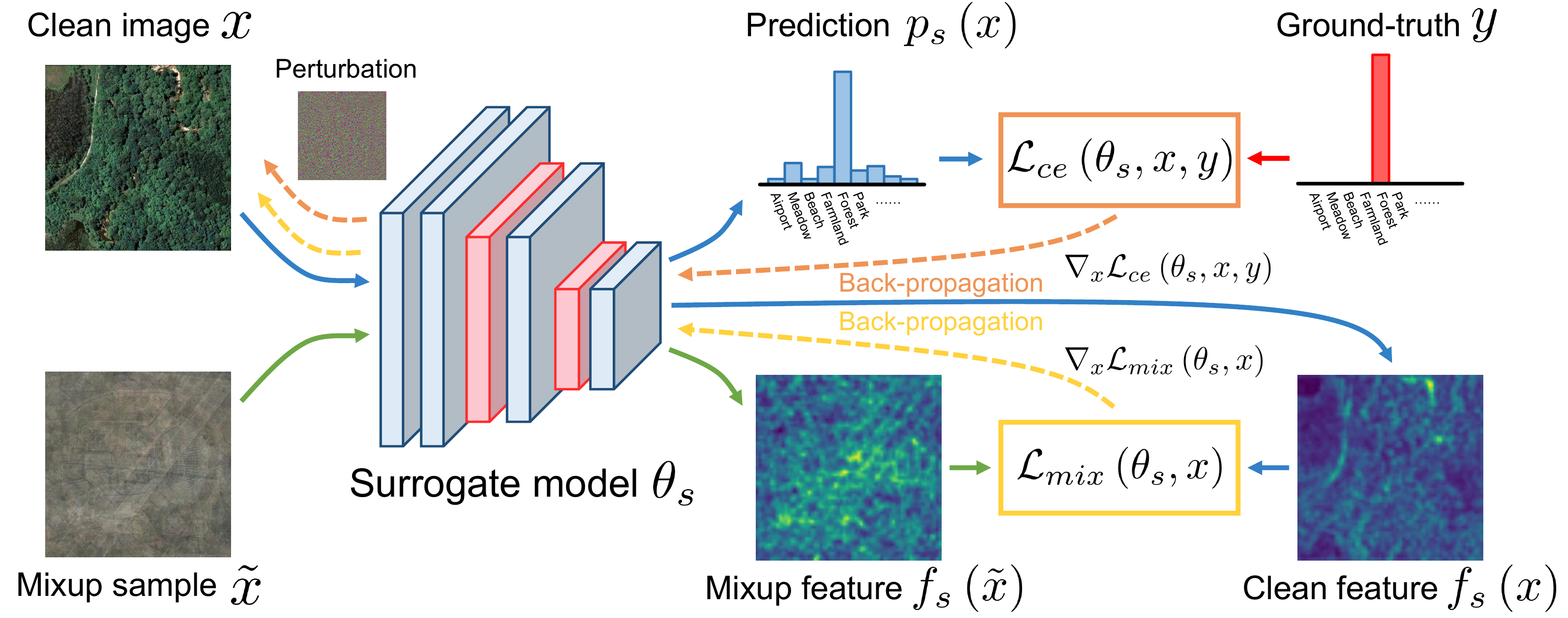}
  \caption{The proposed Mixup-Attack method for an untargeted black-box adversarial attack on a remote sensing scene classification task. The blue and red blocks represent the convolutional layers and the pooling layers in the surrogate model, respectively.}
\label{fig:mixupattack}
\end{figure*}

\section{Methodology}
Most of the previous methods focus on attacking the predicted logits of the surrogate model. The essential idea behind these methods is to change the pixel values of the input image toward the direction that the surrogate model could yield wrong predicted logits. However, due to the discrepancy between different neural networks, the direction that is harmful to the surrogate model may have limited influence on the victim model \cite{akhtar2021advances}. To make the adversarial attack more transferable, we consider attacking the features in the shallow layers of the surrogate model. Since features in the shallow layers generally preserve more detailed spatial information in the image and share similar representations across different networks \cite{yosinski2015understanding}, they may also contain a more common vulnerability. To this end, we first produce a virtual image that belongs to a different category from the input image. Then, instead of directly making the surrogate model yield wrong predicted logits on the input image, we aim to mislead the surrogate model to generate similar feature representations on the input image and the virtual image. Under this circumstance, the generated adversarial examples may also possess stronger transferability to the unknown victim model. Then, a natural question follows: \textit{How can we generate a virtual image whose category can differ from any input image?} To answer this question, this section describes the proposed adversarial attack method in detail.

\subsection{Overview of the Proposed Method}
The flowchart of the proposed Mixup-Attack method is shown in Fig. \ref{fig:mixupattack}. The key idea is to construct a mixup image with the linear combination of training images from different categories and conduct feature-level adversarial attacks on the input image. Specifically, given a surrogate model with parameter $\theta _s$, we extract the shallow features of both the input clean image $x$ and the mixup image $\Tilde{x}$. Then, we define the mix loss function $\mathcal{L}_{mix}$ by minimizing the KL-divergence between features of the mixup image and the input image. Considering the constraint of $\mathcal{L}_{mix}$ does not take the prediction of the network into consideration, the cross-entropy loss $\mathcal{L}_{ce}$ between the predicted logits on the target image and the true label is also adopted to assist the attack. Thus, the complete objective function $\mathcal{L}$ is the weighted combination of $\mathcal{L}_{mix}$ and $\mathcal{L}_{ce}$. Finally, the universal adversarial examples can be produced by adding the gradients (adversarial perturbation) of the complete objective function $\mathcal{L}$ to the original clean image. We further integrate the momentum mechanism into the proposed Mixup-Attack to stabilize the update directions in different iterations.

\subsection{Mixup and Mixcut Samples}
Our initial inspiration for the proposed Mixup-Attack comes from the work in \cite{zhang2017mixup}, where the authors proposed a simple and data-agnostic augmentation routine to construct virtual training examples:
\begin{equation}
\begin{split}
    &\Tilde{x}=\lambda x_1+\left(1-\lambda\right)x_2,\\
    &\Tilde{y}=\lambda y_1+\left(1-\lambda\right)y_2,
\end{split}
\label{eq:mixup}
\end{equation}
where $\left(x_1,y_1\right)$ and $\left(x_2,y_2\right)$ are two samples drawn at random from the training set ($y_1$, and $y_2$ are one-hot label encoding). $\lambda \in \left[0,1\right]$ is a combination ratio for the mixture.

In fact, \eqref{eq:mixup} extends the training distribution by assuming that linear interpolations of feature vectors should lead to linear interpolations of the associated label vectors. Despite its simplicity, Zhang et al. found training with the augmentation in \eqref{eq:mixup} is very useful in improving the classification performance and adversarial robustness of deep neural networks \cite{zhang2017mixup}.

\begin{figure}
  \centering
  \includegraphics[width=\linewidth]{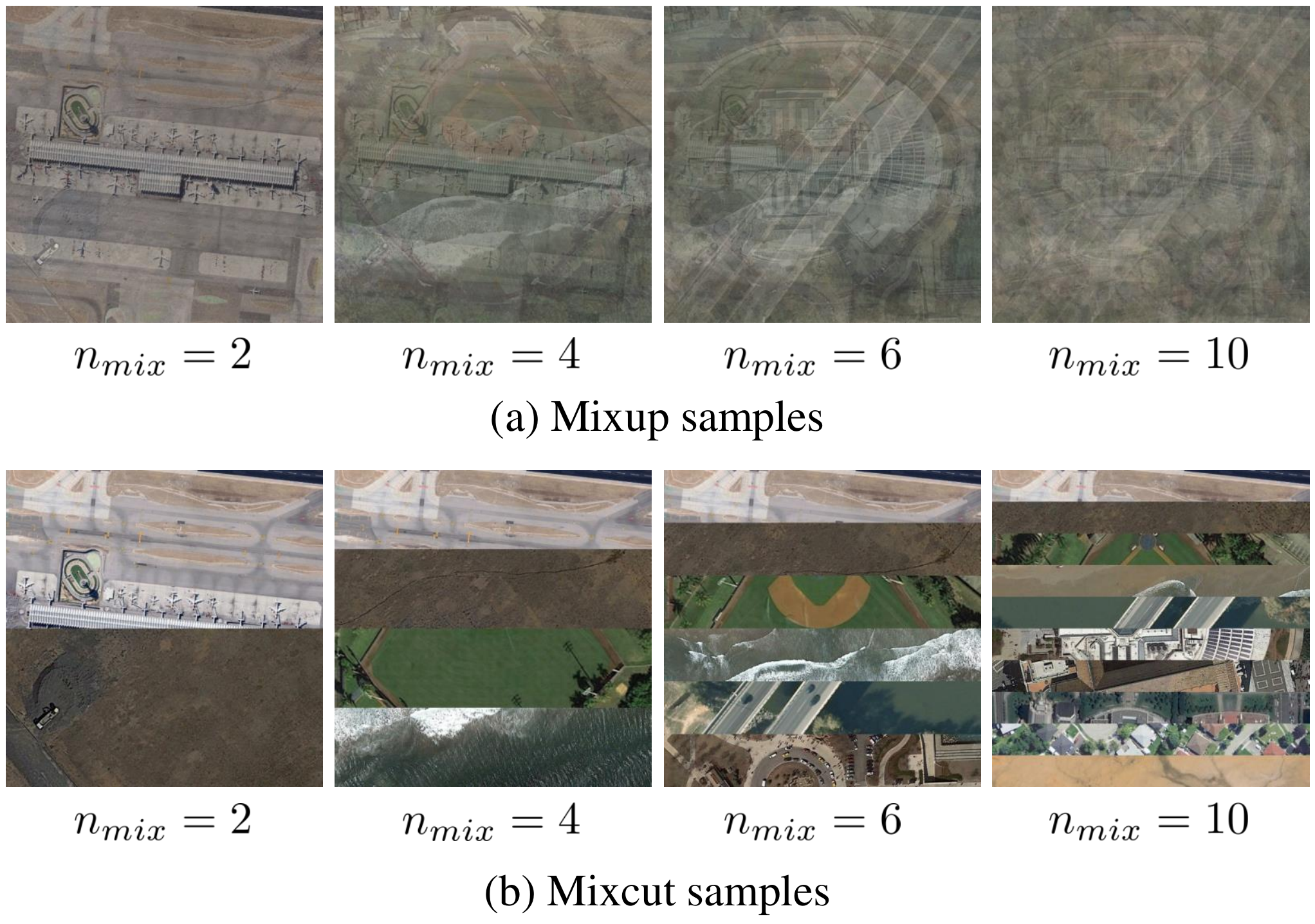}
  \caption{Illustration of the Mixup and Mixcut samples with different values of $n_{mix}$.}
\label{fig:mixup}
\end{figure}

While the original design of \eqref{eq:mixup} is to linearly combine two random samples, we extend this mechanism by involving more images from different categories, considering that our goal is to generate the virtual image $\Tilde{x}$ whose category can differ from any input image:
\begin{equation}
\Tilde{x}=\sum_{i=1}^{n_{mix}}\frac{1}{n_{mix}}x_i,
\label{eq:Mixup}
\end{equation}
where $x_i$ is a random image from the $i$th category in the training set $\left(i=1,2,\cdots,n_{mix}\right)$, and $n_{mix}$ is the number of categories involved in the Mixup-Attack. Fig. \ref{fig:mixup} (a) presents the mixup samples with different values of $n_{mix}$.

Besides the direct linear combination shown in \eqref{eq:mixup}, another possible approach is to stitch the slices from different images. Inspired by the work in \cite{yun2019cutmix}, we further propose the mixcut sampling, which is a simple variant of mixup sampling. Formally, we define the mixcut sampling operation as:
\begin{equation}
\Tilde{x}=\sum_{i=1}^{n_{mix}}M_i\odot x_i,
\label{eq:Mixcut}
\end{equation}
where $M_i\in\{0,1\}^{h\times w}$ denotes a binary mask indicating where to drop out and fill in from the $i$th image $x_i$ $\left(i=1,2,\cdots,n_{mix}\right)$, $h$ and $w$ are the height and width of the image, and $\odot$ is element-wise multiplication. For each binary mask $M_i$, we use the bounding box coordinates $B_i=\left(r_i,c_i,\Delta r_i, \Delta c_i\right)$ to indicate the cropping regions, where $r_i=\left(i-1\right)h/n_{mix}$, and $c_i=0$ are the row and column coordinates of the top-left point in the $i$th bounding box, $\Delta r_i=h/n_{mix}$, and $\Delta c_i=w$ are the height and width of the bounding box. Fig. \ref{fig:mixup} (b) presents the mixcut samples with different values of $n_{mix}$.

Note that here we mainly aim to generate a virtual image whose category can differ from any input image; thus, a larger $n_{mix}$ will generally perform better in the attack since there will be more diverse images from different categories involved. We empirically set $n_{mix}=10$ in our experiments.

\begin{algorithm}[htb]
    \caption{Mixup-Attack/Mixcut-Attack}
    \label{alg:mixattack}
    {\bf Input:}
\begin{enumerate}[-]
\item A surrogate model $\theta _s$ with its feature extraction function $f_s$ and prediction function $p_s$.
\item A clean image $x$ and its ground-truth label $y$.
\item A mixup or mixcut sample generated by \eqref{eq:Mixup} or \eqref{eq:Mixcut}.
\end{enumerate}
 \begin{algorithmic}[1]
    \STATE $g_0\leftarrow 0$; $x_{adv}^0\leftarrow x$.
    \FOR{$t$ in $range\left(0,T\right)$}
    \STATE Compute the mix loss $\mathcal{L}_{mix}\left(\theta _s,x_{adv}^t\right)$, the cross-entropy loss $\mathcal{L}_{ce}\left(\theta _s,x_{adv}^t,y\right)$ via \eqref{eq:mixloss} and \eqref{eq:ce}.
    \STATE Compute the full loss $\mathcal{L}\left(\theta _s,x_{adv}^t,y\right)$ via \eqref{eq:loss}.
    \STATE Update $g_{t+1}$ by accumulating the velocity vector in the gradient direction:\\ \begin{center}$g_{t+1}=g_t+\frac{\nabla _x \mathcal{L}\left(\theta _s,x_{adv}^t,y\right)}{\|\nabla _x \mathcal{L}\left(\theta _s,x_{adv}^t,y\right)\|_{1}}.$\end{center}
    \STATE Update $x_{adv}^{t+1}$ by the normalized  gradients of $g_{t+1}$ with $\ell_{\infty}$ norm:\\ \begin{center}$x_{adv}^{t+1}={\rm clip}\left(x_{adv}^{t}+\alpha \frac{g_{t+1}}{\|g_{t+1}\|_{\infty}}\right).$\end{center}
    \ENDFOR
    \STATE {\bf return} $x_{adv}=x_{adv}^T$.
\end{algorithmic}
\end{algorithm}

\subsection{Feature-Level Attack with Virtual Samples}
Recall that our goal is to produce adversarial examples that share similar feature representation with the virtual sample $\Tilde{x}$ from \eqref{eq:Mixup} or \eqref{eq:Mixcut}. To this end, we propose the mix loss $\mathcal{L}_{mix}$, which minimizes the distribution discrepancy between the features from the adversarial example and the virtual samples. Formally, given a surrogate model with known parameter $\theta _s$, let $f_{s}\left(\cdot\right)$ denote its mapping functions of feature extraction at the shallow layer. Then, the mix loss $\mathcal{L}_{mix}\left(\theta _s,x\right)$ is defined as:
\begin{equation}
\mathcal{L}_{mix}\left(\theta _s,x\right)=-\sum_{r=1}^{n_r}\sum_{c=1}^{n_c}\sum_{k=1}^{n_k} f_{s}\left(x\right)^{(r,c,k)}\log\frac{f_{s}\left(x\right)^{(r,c,k)}}{f_{s}\left(\Tilde{x}\right)^{(r,c,k)}},
\label{eq:mixloss}
\end{equation}
where $n_r,n_c,n_k$ are the numbers of rows, columns, and channels in the feature map, respectively. Note that there is a negative sign in \eqref{eq:mixloss}. Thus, maximizing $\mathcal{L}_{mix}\left(\theta _s,x\right)$ corresponds to minimizing the KL-divergence between features of the mixup image $\Tilde{x}$ and the target image $x$.

Since the constraint in \eqref{eq:mixloss} does not take the prediction of the network into consideration, we further use the cross-entropy loss $\mathcal{L}_{ce}$ to assist the attack. Let $p_{s}\left(\cdot\right)$ denote the mapping functions of the prediction in the surrogate model. Then, the cross-entropy loss $\mathcal{L}_{ce}\left(\theta _s,x,y\right)$ is defined as:
\begin{equation}
\mathcal{L}_{ce}\left(\theta _s,x,y\right)=-\sum_{k=1}^{n_v} y^{(k)}\log p_{s}\left(x\right)^{(k)},
\label{eq:ce}
\end{equation}
where $n_v$ denotes the number of categories in the classification task, and $y$ is the true label of the input sample $x$ with the one-hot encoding.

Finally, the complete loss function of the proposed method is defined as:
\begin{equation}
\mathcal{L}\left(\theta _s,x,y\right)=\mathcal{L}_{mix}\left(\theta _s,x\right)+\beta\mathcal{L}_{ce}\left(\theta _s,x,y\right),
\label{eq:loss}
\end{equation}
where $\beta$ is a weighting parameter for the cross-entropy loss.

With the loss function in \eqref{eq:loss}, the adversarial example $x_{adv}$ can be generated iteratively:
\begin{equation}
x_{adv}^{t+1}={\rm clip}\left(x_{adv}^t+\alpha \frac{\nabla _x \mathcal{L}\left(\theta _s,x_{adv}^t,y\right)}{\|\nabla _x \mathcal{L}\left(\theta _s,x_{adv}^t,y\right)\|_{\infty}} \right),
\label{eq:mixattack}
\end{equation}
where $x_{adv}^{t}$ is the generated adversarial example at the $t$th iteration ($t=0,1,\cdots,T-1$), and $\alpha$ is the step size. When $t=0$, $x_{adv}^0$ is initialized with the input image $x$.

\begin{table*}[]
\centering
\caption{Success rate (\%) of different untargeted adversarial attack methods on the UCM dataset.}
\label{tab:ucm_trans}
\resizebox{\textwidth}{!}{%
\begin{tabular}{ccccccccccc}
\hline
Surrogate & Method & AlexNet & VGG16 & Inception-v3 & ResNet18 & ResNet101 & DenseNet121 & DenseNet201 & RegNetX-400MF & RegNetX-16GF \\
\hline
\multirow{7}{*}{AlexNet} & FGSM \cite{goodfellow2014}& 86.00 & 13.71 & 15.24 & 9.71 & 7.52 & 7.14 & 5.90 & 11.90 & 5.24 \\
 & I-FGSM \cite{kurakin2016adversarial}& \textbf{100} & 34.00 & 31.62 & 31.90 & 22.29 & 21.33 & 18.48 & 35.43 & 13.90 \\
 & C\&W \cite{carlini2017towards}& \textbf{100} & 33.24 & 31.14 & 30.76 & 22.10 & 20.86 & 17.14 & 33.43 & 13.52 \\
 & TPGD \cite{zhang2019theoretically}& 82.67 & 27.90 & 28.95 & 25.62 & 16.86 & 18.57 & 15.81 & 26.10 & 12.10 \\
 & Jitter \cite{schwinn2021exploring}& 96.38 & 13.14 & 16.10 & 9.43 & 7.71 & 6.67 & 6.10 & 10.29 & 5.14 \\
 & Mixup-Attack  (ours) & 78.86 & 55.71 & 56.95 & 49.24 & 43.52 & 31.05 & 28.10 & 47.43 & 40.86 \\
 & Mixcut-Attack (ours) & 89.05 & \textbf{58.29} & \textbf{58.29} & \textbf{52.10} & \textbf{45.71} & \textbf{35.43} & \textbf{31.43} & \textbf{51.33} & \textbf{42.67} \\
\hline
\multirow{7}{*}{ResNet18} & FGSM \cite{goodfellow2014}& 17.43 & 17.81 & 16.00 & 67.33 & 13.05 & 16.67 & 15.33 & 15.90 & 9.43 \\
 & I-FGSM \cite{kurakin2016adversarial}& 42.48 & 48.38 & 31.71 & \textbf{100} & 45.24 & 50.57 & 46.86 & 43.52 & 29.33 \\
 & C\&W \cite{carlini2017towards}& 39.52 & 46.57 & 31.24 & \textbf{100} & 41.33 & 45.90 & 45.62 & 39.81 & 29.05 \\
 & TPGD \cite{zhang2019theoretically}& 37.24 & 39.05 & 25.62 & 75.52 & 33.33 & 39.43 & 33.71 & 31.05 & 20.76 \\
 & Jitter \cite{schwinn2021exploring}& 21.33 & 23.43 & 16.19 & 92.67 & 14.38 & 17.52 & 18.67 & 19.24 & 10.57 \\
 & Mixup-Attack  (ours) & 59.81 & 71.90 & \textbf{70.86} & 83.33 & 49.05 & 67.81 & 52.00 & 50.95 & 54.86 \\
 & Mixcut-Attack (ours) & \textbf{69.14} & \textbf{80.48} & 64.29 & 97.05 & \textbf{71.90} & \textbf{82.86} & \textbf{73.62} & \textbf{72.67} & \textbf{65.14} \\ \hline
\multirow{7}{*}{DenseNet121} & FGSM \cite{goodfellow2014}& 14.00 & 14.29 & 16.48 & 12.67 & 10.38 & 46.48 & 13.14 & 11.62 & 8.00 \\
 & I-FGSM \cite{kurakin2016adversarial}& 39.62 & 44.76 & 30.86 & 55.33 & 41.62 & 97.24 & 57.05 & 40.67 & 35.24 \\
 & C\&W \cite{carlini2017towards}& 36.67 & 41.24 & 29.90 & 53.90 & 38.57 & 97.90 & 55.52 & 39.33 & 32.86 \\
 & TPGD \cite{zhang2019theoretically}& 35.71 & 38.00 & 26.38 & 44.48 & 35.71 & 76.76 & 44.38 & 33.71 & 27.62 \\
 & Jitter \cite{schwinn2021exploring}& 18.95 & 21.62 & 17.14 & 18.19 & 13.71 & 86.00 & 21.05 & 17.52 & 10.86 \\
 & Mixup-Attack  (ours) & \textbf{76.95} & 74.95 & \textbf{78.76} & 72.76 & 70.00 & 93.14 & 73.43 & 70.95 & \textbf{68.76} \\
 & Mixcut-Attack (ours) & 71.14 & \textbf{75.05} & 68.10 & \textbf{78.76} & \textbf{73.05} & \textbf{98.95} & \textbf{84.86} & \textbf{74.29} & 65.14 \\
\hline
\multirow{7}{*}{RegNetX-400MF} & FGSM \cite{goodfellow2014}& 11.90 & 10.67 & 12.57 & 7.52 & 7.24 & 6.86 & 6.48 & 60.10 & 5.43 \\
 & I-FGSM \cite{kurakin2016adversarial}& 21.14 & 22.38 & 18.48 & 19.33 & 14.38 & 17.05 & 15.81 & \textbf{99.71} & 17.14 \\
 & C\&W \cite{carlini2017towards}& 20.57 & 20.76 & 18.95 & 19.24 & 14.10 & 16.19 & 14.29 & 99.62 & 17.14 \\
 & TPGD \cite{zhang2019theoretically}& 19.33 & 19.05 & 16.57 & 15.81 & 11.90 & 14.00 & 12.76 & 84.48 & 13.71 \\
 & Jitter \cite{schwinn2021exploring}& 12.29 & 9.43 & 12.10 & 7.14 & 7.14 & 6.29 & 6.38 & 94.48 & 6.38 \\
 & Mixup-Attack  (ours) & 52.00 & 49.24 & 35.24 & 41.90 & 35.81 & \textbf{41.14} & 37.62 & 85.24 & 34.76 \\
 & Mixcut-Attack (ours) & \textbf{52.38} & \textbf{50.19} & \textbf{36.92} & \textbf{42.10} & \textbf{36.38} & 41.05 & \textbf{39.24} & 86.00 & \textbf{34.86}\\
\hline
\end{tabular}%
}
\\
\vspace{2pt}
\leftline{\scriptsize \quad Note: The leftmost column shows the surrogate models in the adversarial attacks. Best results are highlighted in \textbf{bold}.}
\end{table*}

\subsection{Attack with Momentum}
Inspired by the work in \cite{dong2018boosting}, we further use the momentum strategy to stabilize the update directions at different iterations. Specifically, let $g_t$ denote the momentum term at the $t$th iteration ($t=0,1,\cdots,T-1$). When $t=0$, $g_0$ is initialized as $0$. Then, $g_{t+1}$ is updated by accumulating the velocity vector in the gradient direction as:
\begin{equation}
g_{t+1}=g_t+\frac{\nabla _x \mathcal{L}\left(\theta _s,x_{adv}^t,y\right)}{\|\nabla _x \mathcal{L}\left(\theta _s,x_{adv}^t,y\right)\|_{1}}.
\label{eq:momentum}
\end{equation}

The generated adversarial example $x_{adv}^{t+1}$ at the $\left(t+1\right)$th iteration can thereby be calculated as:
\begin{equation}
x_{adv}^{t+1}={\rm clip}\left(x_{adv}^{t}+\alpha \frac{g_{t+1}}{\|g_{t+1}\|_{\infty}}\right).
\label{eq:mmixattack}
\end{equation}

The detailed implementation procedure of the proposed Mixup-Attack and Mixcut-Attack methods are shown in Algorithm~\ref{alg:mixattack}.

\subsection{Extension to Semantic Segmentation}
In this subsection, we generalize the proposed methods to the semantic segmentation of very high-resolution remote sensing images.

We first randomly select 10 images from the training set to generate the mixup/mixcut sample $\Tilde{x}$ via \eqref{eq:Mixup} or \eqref{eq:Mixcut}. Since semantic segmentation is a pixel-wise classification task, we modify the cross-entropy loss in \eqref{eq:ce} as:
\begin{equation}
\mathcal{L}^{\prime}_{ce}\left(\theta _s,x,y\right)=-\sum_{r=1}^{h}\sum_{c=1}^{w}\sum_{k=1}^{n_v}{ y^{(r,c,k)}\log \rm up}\left(p_{s}\left(x\right)\right)^{(r,c,k)},
\label{eq:ce_seg}
\end{equation}
where $h$ and $w$ are the height and width of the image, and ${\rm up}\left(\cdot\right)$ denotes the upsampling function with the bilinear interpolation.

Accordingly, the complete loss function of the proposed method is modified as:
\begin{equation}
\mathcal{L}\left(\theta _s,x,y\right)=\mathcal{L}_{mix}\left(\theta _s,x\right)+\beta\mathcal{L}^{\prime}_{ce}\left(\theta _s,x,y\right).
\label{eq:loss_seg}
\end{equation}

\begin{table*}[]
\centering
\caption{Success rate (\%) of different untargeted adversarial attack methods on the AID dataset.}
\label{tab:aid_trans}
\resizebox{\textwidth}{!}{%
\begin{tabular}{ccccccccccc}
\hline
Surrogate & Method & AlexNet & VGG16 & Inception-v3 & ResNet18 & ResNet101 & DenseNet121 & DenseNet201 & RegNetX-400MF & RegNetX-16GF \\
\hline
\multirow{7}{*}{AlexNet} & FGSM \cite{goodfellow2014}& 88.74 & 23.32 & 37.94 & 19.40 & 17.74 & 14.78 & 11.92 & 18.62 & 11.46 \\
 & I-FGSM \cite{kurakin2016adversarial}& \textbf{100} & 54.82 & 61.28 & 54.34 & 45.22 & 45.80 & 36.92 & 50.96 & 34.98 \\
 & C\&W \cite{carlini2017towards}& \textbf{100} & 54.04 & 60.34 & 53.12 & 44.80 & 44.94 & 35.88 & 49.60 & 34.58 \\
 & TPGD \cite{zhang2019theoretically}& 81.02 & 44.26 & 59.98 & 42.98 & 37.30 & 36.62 & 27.26 & 39.30 & 27.98 \\
 & Jitter \cite{schwinn2021exploring}& 97.34 & 23.64 & 38.66 & 20.02 & 17.60 & 15.76 & 11.54 & 17.36 & 11.88 \\
 & Mixup-Attack  (ours) & 91.62 & 82.68 & 72.28 & 79.28 & 69.06 & 47.18 & 61.24 & 74.62 & 71.40 \\
 & Mixcut-Attack (ours) & 94.66 & \textbf{84.98} & \textbf{74.56} & \textbf{81.18} & \textbf{71.46} & \textbf{53.14} & \textbf{65.82} & \textbf{77.74} & \textbf{74.48} \\
\hline
\multirow{7}{*}{ResNet18} & FGSM \cite{goodfellow2014}& 22.52 & 28.70 & 38.58 & 82.16 & 29.36 & 31.68 & 24.22 & 24.16 & 22.74 \\
 & I-FGSM \cite{kurakin2016adversarial}& 47.80 & 71.18 & 56.40 & 99.98 & 69.02 & 75.92 & 71.64 & 56.06 & 60.48 \\
 & C\&W \cite{carlini2017towards}& 45.72 & 65.44 & 54.90 & \textbf{100} & 65.04 & 70.50 & 65.42 & 52.90 & 55.72 \\
 & TPGD \cite{zhang2019theoretically}& 40.18 & 60.50 & 49.26 & 75.44 & 55.84 & 59.36 & 54.66 & 42.98 & 48.76 \\
 & Jitter \cite{schwinn2021exploring}& 24.22 & 33.52 & 39.38 & 96.00 & 31.48 & 36.26 & 27.90 & 25.10 & 25.60 \\
 & Mixup-Attack  (ours) & 81.70 & 75.60 & \textbf{77.00} & 94.50 & \textbf{92.08} & 84.86 & 81.64 & 76.34 & \textbf{87.04} \\
 & Mixcut-Attack (ours) & \textbf{81.74} & \textbf{86.54} & 76.52 & 99.98 & 90.26 & \textbf{89.84} & \textbf{90.40} & \textbf{80.82} & 86.66 \\
\hline
\multirow{7}{*}{DenseNet121} & FGSM \cite{goodfellow2014}& 17.14 & 20.20 & 34.92 & 24.68 & 21.72 & 55.04 & 19.56 & 18.06 & 18.24 \\
 & I-FGSM \cite{kurakin2016adversarial}& 41.42 & 63.20 & 51.94 & 74.58 & 62.06 & 99.52 & 73.94 & 51.22 & 56.20 \\
 & C\&W \cite{carlini2017towards}& 40.22 & 59.42 & 50.50 & 69.62 & 60.54 & 99.92 & 70.26 & 48.86 & 53.42 \\
 & TPGD \cite{zhang2019theoretically}& 38.66 & 57.52 & 49.04 & 66.38 & 56.18 & 85.34 & 64.70 & 45.38 & 51.12 \\
 & Jitter \cite{schwinn2021exploring}& 21.88 & 29.16 & 38.86 & 37.00 & 30.48 & 93.46 & 30.66 & 23.36 & 25.38 \\
 & Mixup-Attack  (ours) & \textbf{86.16} & \textbf{88.42} & \textbf{77.60} & 85.84 & \textbf{90.70} & 99.12 & 91.36 & \textbf{85.22} & \textbf{90.24} \\
 & Mixcut-Attack (ours) & 79.88 & 87.38 & 72.78 & \textbf{89.52} & 87.84 & \textbf{99.98} & \textbf{93.88} & 83.40 & 87.08 \\
\hline
\multirow{7}{*}{RegNetX-400MF} & FGSM \cite{goodfellow2014}& 15.12 & 19.10 & 31.76 & 14.52 & 14.68 & 13.28 & 10.58 & 60.50 & 13.46 \\
 & I-FGSM \cite{kurakin2016adversarial}& 27.94 & 44.08 & 43.16 & 36.08 & 38.94 & 34.48 & 28.60 & \textbf{99.96} & 36.04 \\
 & C\&W \cite{carlini2017towards}& 27.06 & 42.62 & 41.54 & 34.20 & 37.26 & 32.30 & 26.44 & 99.82 & 34.44 \\
 & TPGD \cite{zhang2019theoretically}& 24.88 & 41.12 & 40.28 & 30.66 & 34.68 & 28.98 & 22.84 & 80.22 & 31.14 \\
 & Jitter \cite{schwinn2021exploring}& 16.54 & 23.24 & 33.14 & 15.20 & 16.50 & 14.98 & 11.72 & 94.26 & 15.66 \\
 & Mixup-Attack  (ours) & 61.34 & 81.04 & 56.28 & 67.54 & 74.42 & 65.84 & 66.12 & 92.76 & 74.24 \\
 & Mixcut-Attack (ours) & \textbf{61.68} & \textbf{81.30} & \textbf{56.62} & \textbf{68.20} & \textbf{75.08} & \textbf{66.58} & \textbf{66.50} & 92.76 & \textbf{74.80}\\
\hline
\end{tabular}%
}
\\
\vspace{2pt}
\leftline{\scriptsize \quad Note: The leftmost column shows the surrogate models in the adversarial attacks. Best results are highlighted in \textbf{bold}.}
\end{table*}

\section{Experiments}
\subsection{Data Descriptions}
\subsubsection{Scene Classification}
Two benchmark remote sensing image datasets for scene classification, the UC
Merced (UCM)\footnote{http://weegee.vision.ucmerced.edu/datasets/landuse.html} \cite{yang2010bag} and the AID\footnote{https://captain-whu.github.io/AID/} \cite{aid}, are utilized in this study.

\textbf{UCM} consists of 2100 overhead scene images with 21 land-use classes. Each class contains 100 aerial images measuring 256$\times$256 pixels, with a spatial resolution of 0.3 m per pixel in the red-green-blue color space. This dataset is extracted from aerial orthoimagery downloaded from the U.S. Geological Survey (USGS) National Map. The 21 land-use classes are: agricultural, airplane, baseball diamond, beach, buildings, chaparral, dense residential, forest, freeway, golf course, harbor, intersection, medium-density residential, mobile home park, overpass, parking lot, river, runway, sparse residential, storage tanks, and tennis courts.

\textbf{AID} is collected from Google Earth (Google Inc.). It is made up of the following 30 aerial scene types: airport, bare land, baseball field, beach, bridge, center, church, commercial, dense residential, desert, farmland, forest, industrial, meadow, medium residential, mountain, park, parking, playground, pond, port, railway station, resort, river, school, sparse residential, square, stadium, storage tanks, and viaduct. All the images are labeled by specialists in the field of remote sensing image interpretation. The numbers of sample images vary a lot with different aerial scene types, from 220 up to 420. In all, the AID dataset has a number of 10,000 images within 30 classes. The AID dataset has multiple resolutions: the pixel resolution changes from about 8 m to about 0.5 m. The size of each aerial image is fixed at 600$\times$600 pixels.
\subsubsection{Semantic Segmentation}
Two benchmark very high-resolution remote sensing image datasets for semantic segmentation, the Vaihingen\footnote{http://www2.isprs.org/commissions/comm3/wg4/2d-sem-label-vaihingen.
html} \cite{cramer2010dgpf} and the Zurich Summer\footnote{https://sites.google.com/site/michelevolpiresearch/data/zurich-dataset} \cite{volpi2015semantic}, are utilized in this study.

\textbf{Vaihingen} is a benchmark dataset for semantic segmentation provided by the International Society for Photogrammetry and Remote Sensing (ISPRS); it is a subset of the data used by the German Association of Photogrammetry and Remote Sensing (DGPF) to test digital aerial cameras \cite{cramer2010dgpf}. There is a total of $33$ aerial images with a spatial resolution of $9$ cm collected over the city of Vaihingen. The average size of an image is around $2500\times 1900$ pixels, covering an area of about $1.38$ km$^2$. For each aerial image, three bands are available: the near-infrared, red, and green. Among these images, $16$ of them are fully annotated with $6$ different land-cover classes: impervious surface, building, low vegetation, tree, car, and clutter/background. We use the same train-test split protocol as specified in the previous work \cite{hua2021semantic} and select five images (image IDs: 11, 15, 28, 30, 34) as the test set. The remaining images are utilized to make up the training set.

\begin{figure}
  \centering
  \includegraphics[width=\linewidth]{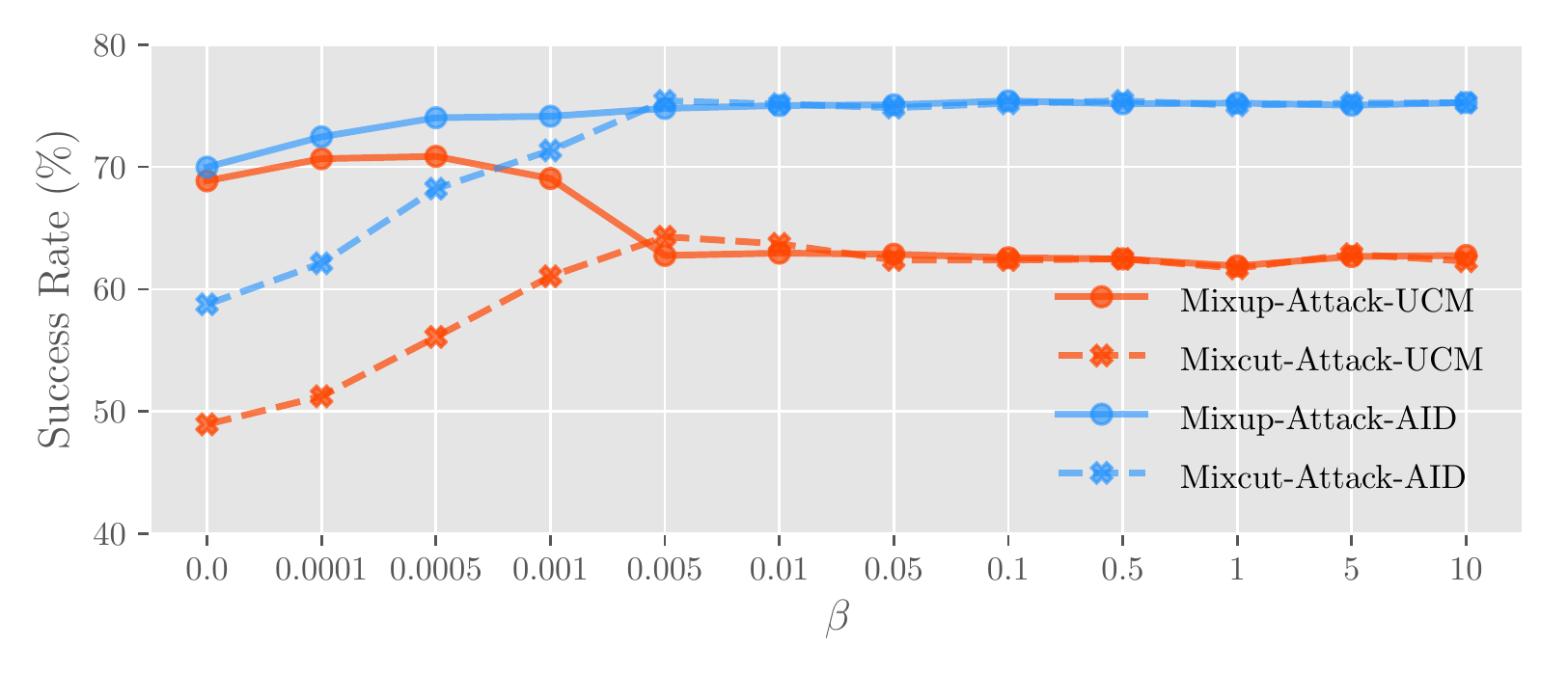}
  \caption{Success rate (\%) of the proposed Mixup-Attack and Mixcut-Attack from ResNet18$\rightarrow$Inception-v3 with different values of $\beta$ on UCM and AID datasets.}
\label{fig:weight}
\end{figure}

\begin{table}
\centering
\caption{Performance contribution of each module in Mixup-Attack and Mixcut-Attack on the scene classification task.}
\label{tab:ablation_cls}
\begin{tabular*}{\linewidth}{@{\extracolsep{\fill}}ccccc}
\hline
\multicolumn{5}{c}{Success Rate (\%) from ResNet18$\rightarrow$Inception-v3} \\ \hline
$\mathcal{L}_{ce}$ & $\mathcal{L}_{mix}$ &Momentum &UCM & AID \\ \hline
\textit{Mixup-Attack} &  &  &  &  \\
$\checkmark$ &  &  & 31.71 & 56.40 \\
$\checkmark$ & $\checkmark$ &  & 64.57 & 64.54 \\
$\checkmark$ & $\checkmark$ & $\checkmark$ & \textbf{70.86} & \textbf{77.00} \\ \hline
\textit{Mixcut-Attack} &  &  &  &  \\
$\checkmark$ &  &  & 31.71 & 56.40 \\
$\checkmark$ & $\checkmark$ &  & 43.05 & 61.28 \\
$\checkmark$ & $\checkmark$ & $\checkmark$ & \textbf{64.29} & \textbf{76.52} \\ \hline
\end{tabular*}%
\\
\vspace{2pt}
\leftline{\scriptsize Note: Results are reported in success rate (\%). Best results are highlighted in \textbf{bold}.}
\end{table}

\textbf{Zurich Summer} consists of $20$ satellite images, taken over the city of Zurich in August 2002 by the Quick-Bird satellite \cite{volpi2015semantic}. The spatial resolution is $0.62$ m, and the average size of images is around $1000\times 1000$ pixels. The images consist of four channels: the near-infrared, red, green, and blue. Similar to previous work \cite{hua2021semantic}, we only utilize the near-infrared, red, and green channels in the experiments and select five images (image IDs: 16, 17, 18, 19, 20) as the test set. The remaining $15$ images are utilized to make up the training set. In total, there are $8$ urban classes: road, building, tree, grass, bare soil, water, railway, and swimming pool. Uncategorized pixels are labeled as background.

\begin{table*}
\centering
\caption{Success rate (\%) of different untargeted adversarial attack methods on the Vaihingen dataset.}
\label{tab:vaihingen_trans}
\resizebox{\textwidth}{!}{%
\begin{tabular*}{1.2\linewidth}{@{\extracolsep{\fill}}ccccccccccc}
 \hline
Surrogate & Method & FCN-32s & FCN-8s & DeepLab-v2 & U-Net & SegNet & PSPNet & SQNet & LinkNet & FRRNet-A \\ \hline
\multirow{7}{*}{FCN-8s} & FGSM \cite{goodfellow2014}& 24.86 & 22.09 & 19.17 & 18.87 & 20.29 & 17.88 & 17.02 & 17.69 & 17.31 \\
 & I-FGSM \cite{kurakin2016adversarial}& 32.39 & 36.65 & 24.37 & 27.59 & 24.57 & 25.63 & 20.49 & 23.37 & 25.17 \\
 & C\&W \cite{carlini2017towards}& 42.72 & 64.28 & 30.77 & 38.10 & 31.62 & 34.06 & 24.68 & 28.60 & 34.78 \\
 & TPGD \cite{zhang2019theoretically}& 26.35 & 24.41 & 20.10 & 20.54 & 20.80 & 19.35 & 17.28 & 18.56 & 19.04 \\
 & Jitter \cite{schwinn2021exploring}& 25.27 & 22.55 & 19.33 & 19.02 & 20.39 & 18.07 & 17.15 & 17.86 & 17.52 \\
 & Mixcut-Attack (ours) & 37.35 & 49.65 & 32.77 & 47.37 & 44.31 & 41.55 & 35.27 & 41.14 & 48.26 \\
 & Mixup-Attack  (ours) & \textbf{54.21} & \textbf{65.97} & \textbf{53.92} & \textbf{71.11} & \textbf{71.64} & \textbf{68.85} & \textbf{52.11} & \textbf{65.23} & \textbf{72.05} \\ \hline
\multirow{7}{*}{U-Net} & FGSM \cite{goodfellow2014}& 24.40 & 18.41 & 18.71 & 23.13 & 20.17 & 17.88 & 16.72 & 17.80 & 17.17 \\
 & I-FGSM \cite{kurakin2016adversarial}& 27.00 & 21.55 & 20.44 & 41.82 & 22.99 & 22.83 & 18.45 & 20.88 & 22.80 \\
 & C\&W \cite{carlini2017towards}& 28.06 & 22.46 & 20.94 & \textbf{62.85} & 24.80 & 23.91 & 19.09 & 21.13 & 26.47 \\
 & TPGD \cite{zhang2019theoretically}& 25.40 & 19.30 & 19.30 & 30.90 & 20.76 & 18.77 & 17.05 & 18.56 & 19.06 \\
 & Jitter \cite{schwinn2021exploring}& 24.55 & 18.48 & 18.73 & 24.09 & 20.17 & 18.02 & 16.83 & 17.93 & 17.43 \\
 & Mixcut-Attack (ours) & 31.68 & 27.48 & 23.80 & 47.74 & 28.83 & 29.23 & 25.25 & 26.05 & 29.97 \\
 & Mixup-Attack  (ours) & \textbf{46.31} & \textbf{49.41} & \textbf{35.72} & 58.78 & \textbf{47.17} & \textbf{47.82} & \textbf{47.81} & \textbf{45.81} & \textbf{50.27} \\ \hline
\multirow{7}{*}{PSPNet} & FGSM \cite{goodfellow2014}& 24.29 & 18.29 & 18.60 & 18.57 & 19.97 & 19.78 & 16.64 & 17.63 & 16.81 \\
 & I-FGSM \cite{kurakin2016adversarial}& 26.47 & 20.54 & 19.86 & 23.04 & 21.16 & 32.14 & 18.07 & 20.45 & 20.12 \\
 & C\&W \cite{carlini2017towards}& 27.79 & 21.25 & 20.20 & 24.80 & 22.02 & 42.83 & 18.58 & 21.11 & 21.54 \\
 & TPGD \cite{zhang2019theoretically}& 25.13 & 18.80 & 19.00 & 19.77 & 19.95 & 24.27 & 16.79 & 18.33 & 17.52 \\
 & Jitter \cite{schwinn2021exploring}& 24.33 & 18.25 & 18.63 & 18.61 & 19.91 & 20.43 & 16.65 & 17.72 & 16.93 \\
 & Mixcut-Attack (ours) & 28.54 & 22.90 & 21.20 & 26.02 & 23.31 & 34.47 & 21.07 & 23.18 & 23.03 \\
 & Mixup-Attack  (ours) & \textbf{41.10} & \textbf{43.75} & \textbf{31.75} & \textbf{48.82} & \textbf{45.00} & \textbf{53.41} & \textbf{43.31} & \textbf{44.87} & \textbf{45.86} \\ \hline
\multirow{7}{*}{LinkNet} & FGSM \cite{goodfellow2014}& 24.79 & 19.00 & 19.03 & 19.54 & 20.46 & 18.64 & 17.59 & 21.86 & 17.68 \\
 & I-FGSM \cite{kurakin2016adversarial}& 28.35 & 23.50 & 21.86 & 26.51 & 24.15 & 25.98 & 21.90 & 38.23 & 24.23 \\
 & C\&W \cite{carlini2017towards}& 30.38 & 25.15 & 22.85 & 29.87 & 27.92 & 29.53 & 24.28 & 55.37 & 28.14 \\
 & TPGD \cite{zhang2019theoretically}& 26.12 & 20.64 & 19.94 & 22.70 & 21.36 & 21.55 & 18.88 & 29.64 & 20.53 \\
 & Jitter \cite{schwinn2021exploring}& 24.78 & 18.95 & 19.02 & 19.49 & 20.46 & 18.66 & 17.50 & 22.32 & 17.69 \\
 & Mixcut-Attack (ours) & 42.34 & 40.11 & 33.92 & 42.67 & 41.88 & 41.75 & 42.72 & 52.01 & 41.14 \\
 & Mixup-Attack  (ours) & \textbf{50.74} & \textbf{52.60} & \textbf{40.83} & \textbf{53.56} & \textbf{51.95} & \textbf{47.16} & \textbf{55.62} & \textbf{55.71} & \textbf{49.61} \\ \hline
\end{tabular*}}%
\\
\vspace{2pt}
\leftline{\scriptsize \quad Note: The leftmost column shows the surrogate models in the adversarial attacks. Best results are highlighted in \textbf{bold}.}
\end{table*}

\begin{table*}
\centering
\caption{Success rate (\%) of different untargeted adversarial attack methods on the Zurich Summer dataset.}
\label{tab:zurich_trans}
\resizebox{\textwidth}{!}{%
\begin{tabular*}{1.2\linewidth}{@{\extracolsep{\fill}}ccccccccccc}
 \hline
Surrogate & Method & FCN-32s & FCN-8s & DeepLab-v2 & U-Net & SegNet & PSPNet & SQNet & LinkNet & FRRNet-A \\ \hline
\multirow{7}{*}{FCN-8s} & FGSM \cite{goodfellow2014}& 25.81 & 13.60 & 11.75 & 10.91 & 13.46 & 11.71 & 12.00 & 10.74 & 8.88 \\
 & I-FGSM \cite{kurakin2016adversarial}& 31.54 & 25.10 & 19.53 & 17.57 & 18.11 & 16.75 & 16.02 & 14.98 & 15.03 \\
 & C\&W \cite{carlini2017towards}& 43.11 & \textbf{60.56} & 33.06 & 28.00 & 28.78 & 22.57 & 22.58 & 24.27 & 22.12 \\
 & TPGD \cite{zhang2019theoretically}& 27.86 & 16.33 & 14.01 & 13.03 & 15.75 & 13.13 & 12.39 & 10.91 & 11.95 \\
 & Jitter \cite{schwinn2021exploring}& 27.40 & 13.61 & 11.45 & 11.75 & 14.52 & 11.55 & 12.09 & 9.38 & 9.49 \\
 & Mixcut-Attack (ours) & 47.28 & 39.83 & 31.25 & 24.10 & 45.15 & 24.15 & 27.19 & 26.56 & 22.53 \\
 & Mixup-Attack  (ours) & \textbf{52.88} & 46.55 & \textbf{36.66} & \textbf{31.93} & \textbf{52.37} & \textbf{28.29} & \textbf{35.18} & \textbf{33.57} & \textbf{26.32} \\ \hline
\multirow{7}{*}{U-Net} & FGSM \cite{goodfellow2014}& 25.48 & 11.05 & 10.57 & 13.59 & 13.69 & 11.24 & 11.51 & 10.59 & 9.18 \\
 & I-FGSM \cite{kurakin2016adversarial}& 27.94 & 15.28 & 13.95 & 25.69 & 17.46 & 15.04 & 14.51 & 13.45 & 15.17 \\
 & C\&W \cite{carlini2017towards}& 30.84 & 18.00 & 16.68 & 44.91 & 22.87 & 16.68 & 16.63 & 15.60 & 17.03 \\
 & TPGD \cite{zhang2019theoretically}& 26.98 & 11.64 & 11.42 & 18.84 & 15.60 & 11.93 & 12.24 & 10.10 & 12.11 \\
 & Jitter \cite{schwinn2021exploring}& 26.65 & 10.78 & 10.56 & 13.45 & 14.23 & 11.21 & 11.77 & 9.06 & 9.48 \\
 & Mixcut-Attack (ours) & 30.38 & 18.19 & 18.00 & 28.99 & 21.85 & 19.34 & 17.68 & 19.51 & 19.23 \\
 & Mixup-Attack  (ours) & \textbf{37.26} & \textbf{30.98} & \textbf{24.43} & \textbf{52.41} & \textbf{34.42} & \textbf{30.17} & \textbf{31.50} & \textbf{31.70} & \textbf{38.42} \\ \hline
\multirow{7}{*}{PSPNet} & FGSM \cite{goodfellow2014}& 27.06 & 11.34 & 10.95 & 12.19 & 14.60 & 13.44 & 12.50 & 9.68 & 10.06 \\
 & I-FGSM \cite{kurakin2016adversarial}& 28.68 & 14.62 & 13.50 & 17.81 & 17.80 & 26.81 & 15.94 & 14.72 & 16.12 \\
 & C\&W \cite{carlini2017towards}& 29.61 & 17.00 & 16.63 & 20.07 & 18.90 & 38.14 & 18.45 & 17.99 & 15.18 \\
 & TPGD \cite{zhang2019theoretically}& 26.27 & 11.50 & 11.16 & 11.92 & 14.88 & 13.62 & 11.70 & 9.99 & 9.95 \\
 & Jitter \cite{schwinn2021exploring}& 26.65 & 10.70 & 10.51 & 11.24 & 14.06 & 12.16 & 11.84 & 9.08 & 9.15 \\
 & Mixcut-Attack (ours) & 29.77 & 17.83 & 16.49 & 22.44 & 19.61 & 26.20 & 17.12 & 17.92 & 17.54 \\
 & Mixup-Attack  (ours) & \textbf{40.96} & \textbf{34.44} & \textbf{25.51} & \textbf{56.62} & \textbf{33.18} & \textbf{39.13} & \textbf{34.74} & \textbf{38.46} & \textbf{30.76} \\ \hline
\multirow{7}{*}{LinkNet} & FGSM \cite{goodfellow2014}& 25.48 & 11.60 & 12.72 & 10.93 & 14.33 & 13.49 & 12.77 & 14.23 & 10.40 \\
 & I-FGSM \cite{kurakin2016adversarial}& 30.44 & 19.16 & 19.16 & 19.14 & 21.46 & 21.61 & 20.72 & 30.64 & 18.25 \\
 & C\&W \cite{carlini2017towards}& 33.67 & 23.14 & 22.14 & 25.77 & 26.46 & 26.32 & 25.05 & \textbf{48.94} & 21.62 \\
 & TPGD \cite{zhang2019theoretically}& 26.61 & 13.16 & 13.34 & 13.47 & 15.24 & 13.90 & 13.13 & 15.64 & 12.43 \\
 & Jitter \cite{schwinn2021exploring}& 26.74 & 11.12 & 10.89 & 11.96 & 14.75 & 12.84 & 12.51 & 12.21 & 10.91 \\
 & Mixcut-Attack (ours) & 37.27 & 25.56 & 22.94 & 32.77 & 39.52 & 31.95 & 30.29 & 37.52 & 25.32 \\
 & Mixup-Attack  (ours) & \textbf{43.36} & \textbf{30.27} & \textbf{27.66} & \textbf{43.45} & \textbf{47.23} & \textbf{38.78} & \textbf{35.92} & 40.61 & \textbf{29.54}\\ \hline
\end{tabular*}%
}
\\
\vspace{2pt}
\leftline{\scriptsize \quad Note: The leftmost column shows the surrogate models in the adversarial attacks. Best results are highlighted in \textbf{bold}.}
\end{table*}

\subsection{Experimental Settings and Implementation Details}
We adopt FGSM \cite{goodfellow2014}, I-FGSM \cite{kurakin2016adversarial}, C\&W \cite{carlini2017towards}, TPGD \cite{zhang2019theoretically}, and Jitter \cite{schwinn2021exploring} as the comparison methods, along with the proposed Mixup-Attack and Mixcut-Attack methods, to conduct the untargeted black-box adversarial attack for both scene classification and semantic segmentation tasks. The perturbation level $\epsilon$ and the step size $\alpha$ in all methods are fixed to $1$. For iterative methods like I-FGSM, C\&W, TPGD, Jitter, and the proposed methods, we fix the number of total iterations $T$ to $5$. All methods in this study use the $\ell_{\infty}$ norm for the calculation of adversarial perturbation.

The feature extraction function $f_s\left(\cdot\right)$ in \eqref{eq:mixloss} is implemented with the first pooling layer in each surrogate model. The weighting factor $\beta$ in \eqref{eq:loss} and \eqref{eq:loss_seg} is set as $0.0005$ for Mixup-Attack and $0.005$ for Mixcut-Attack (see Fig. \ref{fig:weight} for the parameter analysis). Scale augmentation \cite{lin2019nesterov} is adopted to improve the generalization ability of the model.

\begin{figure*}
  \centering
  \includegraphics[width=0.95\linewidth]{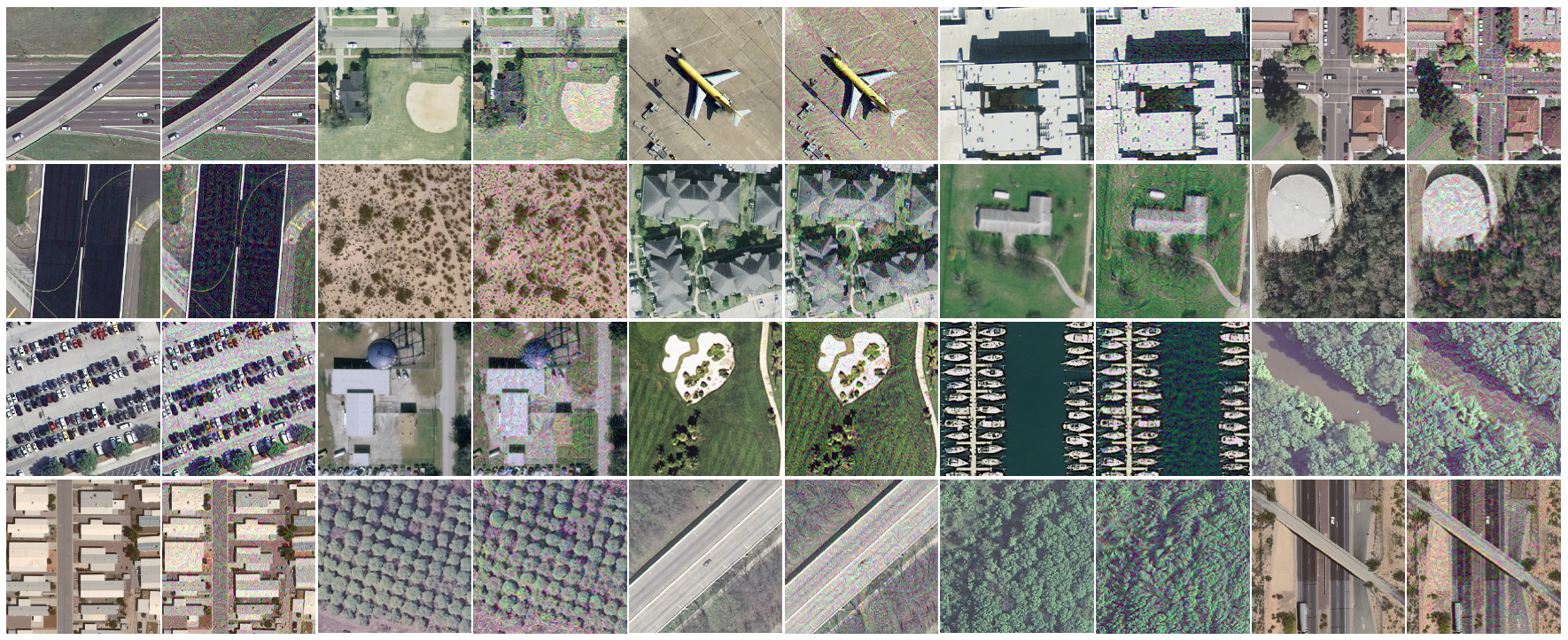}
  \caption{Example images in the UCM dataset (left image in each pair) and the corresponding adversarial examples generated by the proposed Mixcut-Attack method with ResNet18 in the UAE-RS dataset (right image in each pair).}
\label{fig:ucm}
\end{figure*}

\begin{figure*}
  \centering
  \includegraphics[width=0.95\linewidth]{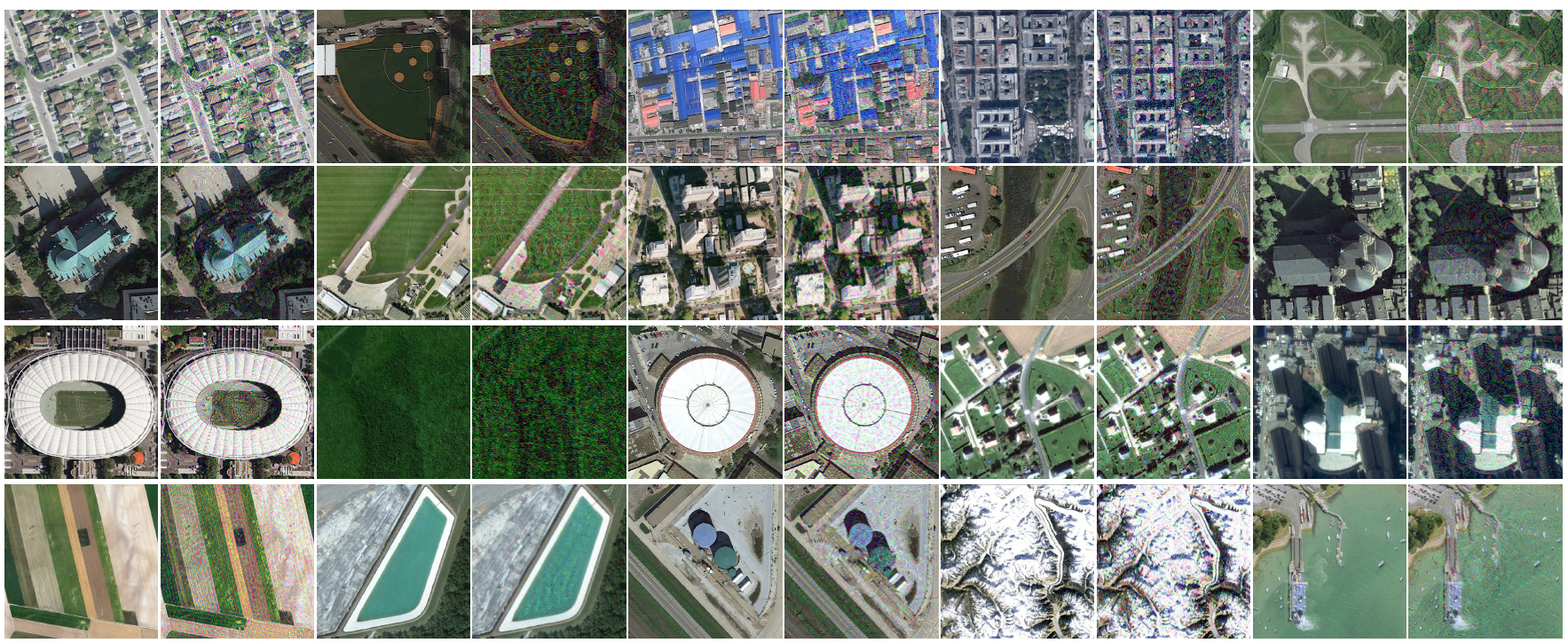}
  \caption{Example images in the AID dataset (left image in each pair) and the corresponding adversarial examples generated by the proposed Mixcut-Attack method with ResNet18 in the UAE-RS dataset (right image in each pair).}
\label{fig:aid}
\end{figure*}

\begin{table}[]
\centering
\caption{Performance contribution of each module in Mixup-Attack and Mixcut-Attack on the semantic segmentation task.}
\label{tab:ablation_seg}
\begin{tabular*}{\linewidth}{@{\extracolsep{\fill}}ccccc}
\hline
\multicolumn{5}{c}{Success Rate (\%) from FCN8s$\rightarrow$SegNet} \\ \hline
$\mathcal{L}_{ce}$ & $\mathcal{L}_{mix}$ & Momentum & Vaihingen & Zurich Summer \\ \hline
\textit{Mixcut-Attack} &  &  &  &  \\
$\checkmark$ &  &  & 24.57 & 18.11 \\
$\checkmark$ & $\checkmark$ &  & 40.92 & 44.68 \\
$\checkmark$ & $\checkmark$ & $\checkmark$ & \textbf{44.31} & \textbf{45.15}\\ \hline
\textit{Mixup-Attack} &  &  &  &  \\
$\checkmark$ &  & & 24.57   & 18.11\\
$\checkmark$ & $\checkmark$ &  & 70.20 & 52.16 \\
$\checkmark$ & $\checkmark$ & $\checkmark$ & \textbf{71.64} & \textbf{52.37} \\ \hline
\end{tabular*}%
\\
\vspace{2pt}
\leftline{\scriptsize Note: Results are reported in success rate (\%). Best results are highlighted in \textbf{bold}.}
\end{table}

For the scene classification task, we randomly select $50\%$ of the samples for the training set and use the remainder for the test set. We use different surrogate models trained on the training set to generate adversarial examples on the test set with the aforementioned methods, which are then fed to different deep neural networks to evaluate the attack's performance. We adopt the success rate $SR=n_{wrong}/n_{total}$ as the evaluation metric, where $n_{wrong}$ denotes the number of misclassified samples and $n_{total}$ is the number of samples in the test set. A higher success rate indicates that the generated adversarial examples possess stronger transferability to the target model. For the semantic segmentation task, we calculate $n_{wrong}$ by counting the number of all misclassified pixels in the test set, while $n_{total}$ is the sum of all valid pixels in the test set.

The experiments in this study are implemented with the PyTorch platform \cite{paszke2019pytorch} using two NVIDIA Tesla V100 (32GB) GPUs.

\begin{figure*}
  \centering
  \includegraphics[width=\linewidth]{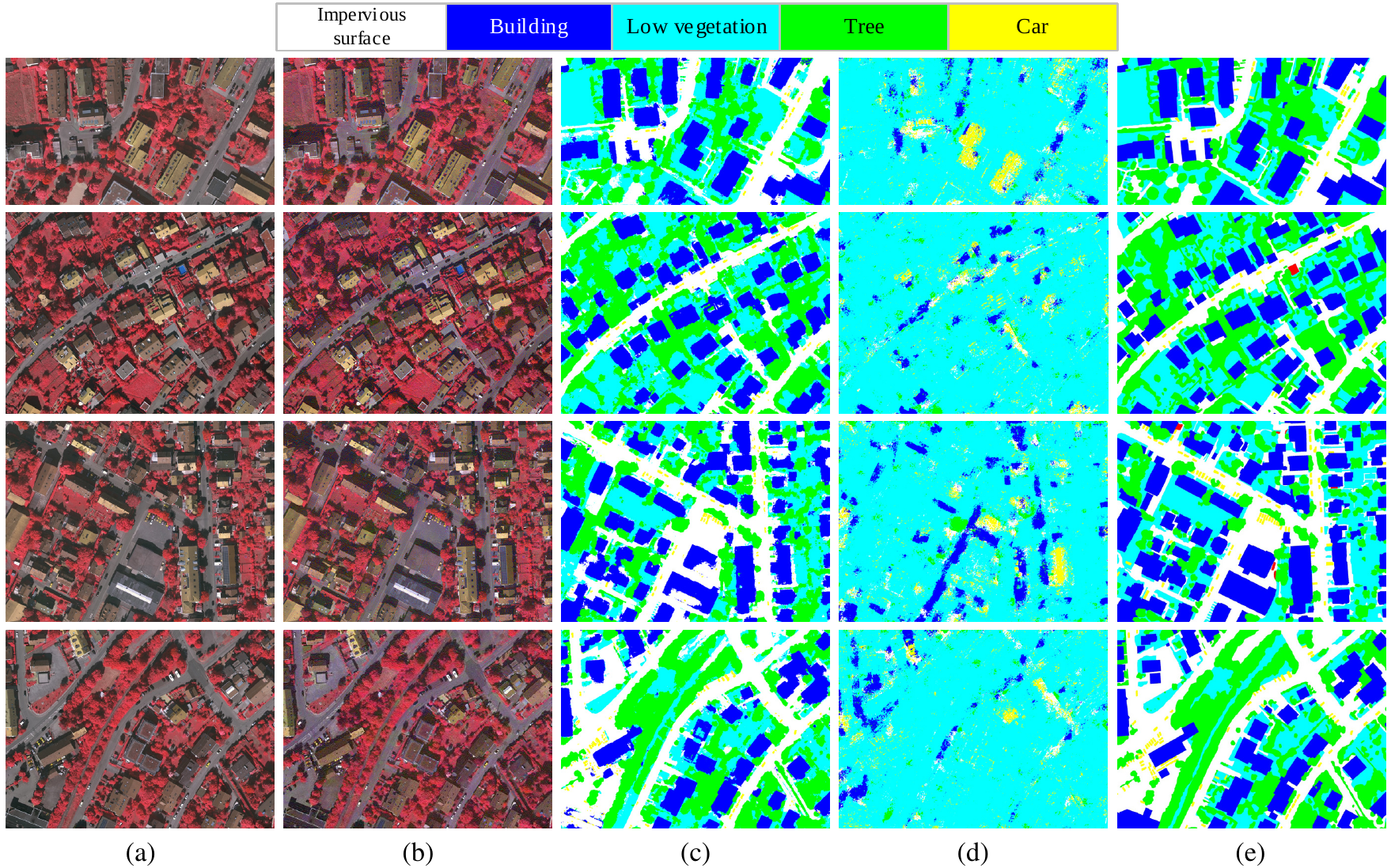}
  \caption{Qualitative results of the black-box adversarial attacks from FCN-8s $\rightarrow$ SegNet on the Vaihingen dataset using the proposed Mixup-Attack method. (a) The original clean test images in the Vaihingen dataset. (b) The corresponding adversarial examples in the UAE-RS dataset. (c)  Segmentation results of SegNet on the clean images. (d) Segmentation results of SegNet on the adversarial images. (e) Ground-truth annotations.}
\label{fig:vaihingen}
\end{figure*}

\subsection{Black-Box Attacks on Scene Classification}
Four surrogate models, AlexNet \cite{alexnet}, ResNet18 \cite{resnet}, DenseNet121 \cite{densenet}, and RegNetX-400MF \cite{regnet}, are adopted to evaluate the performance of different adversarial attack methods on scene classification.

The detailed quantitative results are presented in Tables \ref{tab:ucm_trans} and \ref{tab:aid_trans}. It can be observed that in all black-box attack scenarios where the target model is different from the surrogate model, the proposed Mixup-Attack and Mixcut-Attack can outperform the comparison methods by a large margin in both datasets. Take the results of ResNet18$\rightarrow$Inception-v3 on the UCM dataset, for example. While C\&W obtains a success rate of about $31\%$, both Mixup-Attack and Mixcut-Attack can achieve a success rate of over $64\%$, which is much higher with more than $30$ percentage points. Similar phenomena can be observed in the AID dataset. These results indicate that the proposed methods can generate adversarial examples with stronger transferability to different target models.

It can also be observed that the white-box adversarial attack performance of the proposed methods is generally lower than that of I-FGSM or C\&W methods. Take the results of ResNet18$\rightarrow$ResNet18 in the UCM dataset, for example. While both I-FGSM and C\&W methods obtain a success rate of $100\%$, Mixup-Attack yields a success rate of about $83\%$ in this case. One intuitive reason for this phenomenon is that the mix loss $\mathcal{L}_{mix}\left(\theta _s,x\right)$ used in the proposed methods may sacrifice a little bit of white-box attack performance to achieve better transferability, as it does not directly mislead the network to yield wrong predicted logits. Nevertheless, in most cases, the proposed methods can still obtain a success rate over $80\%$ in both datasets, which would be a serious threat in the white-box adversarial attack scenario.

The parameter $\beta$ in \eqref{eq:loss} is an important factor in the proposed methods. Fig. \ref{fig:weight} shows the success rate of the proposed Mixup-Attack and Mixcut-Attack from ResNet18$\rightarrow$Inception-v3 with different values of $\beta$ on UCM and AID datasets. It can be observed that for Mixup-Attack, a relatively small $\beta$ would be more beneficial to the performance, especially on the UCM dataset, where the highest success rate is obtained with $\beta=0.0005$. For Mixcut-Attack, it can be observed that the $\mathcal{L}_{ce}\left(\theta _s,x,y\right)$ term plays an important role in the model, as the success rate increases accordingly as the value of $\beta$ grows on both datasets. When $\beta>0.005$, the success rate gradually becomes saturated. Based on the above analysis, we set $\beta=0.0005$ for Mixup-Attack and $\beta=0.005$ for Mixcut-Attack in the experiments.

To evaluate how each module in the proposed methods would influence the adversarial attack performance, the quantitative ablation study results are presented in Table \ref{tab:ablation_cls}. In both UCM and AID datasets, we find that directly using the cross-entropy loss $\mathcal{L}_{ce}$ alone only leads to a limited success rate, while combining $\mathcal{L}_{ce}$ and $\mathcal{L}_{mix}$ can significantly improve the performance by around $33$ and $8$ percentage points for Mixup-Attack, and $12$ and $5$ percentage points for Mixcut-Attack. Finally, with the help of the momentum strategy, the success rate is further increased, achieving state-of-the-art performance.

\begin{figure*}
  \centering
  \includegraphics[width=\linewidth]{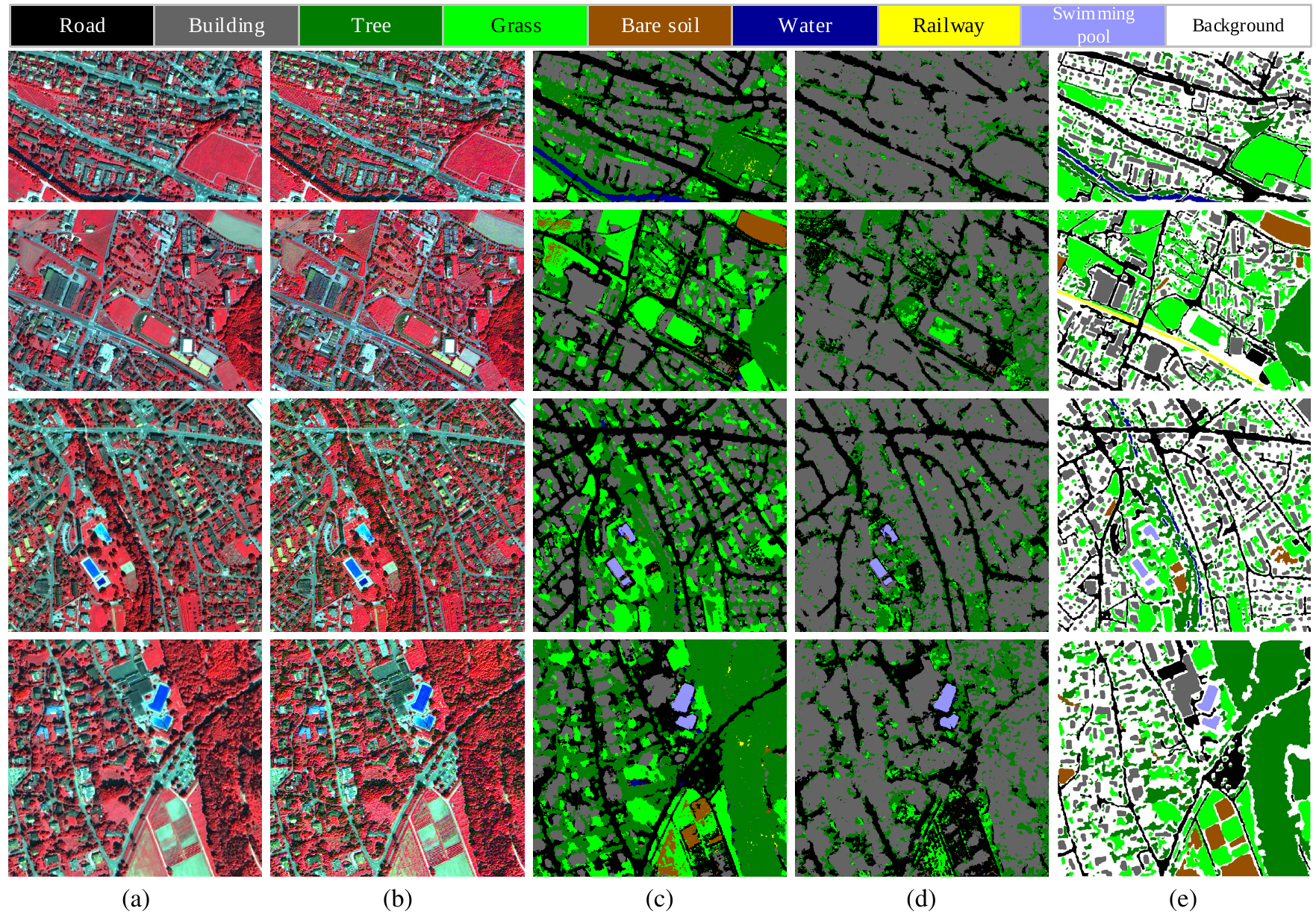}
  \caption{Qualitative results of the black-box adversarial attacks from FCN-8s $\rightarrow$ SegNet on the Zurich Summer dataset using the proposed Mixup-Attack method. (a) The original clean test images in the Zurich dataset. (b) The corresponding adversarial examples in the UAE-RS dataset. (c) Segmentation results of SegNet on the clean images. (d) Segmentation results of SegNet on the adversarial images. (e) Ground-truth annotations.}
\label{fig:zurich}
\end{figure*}

\subsection{Black-Box Attacks on Semantic Segmentation}
Four surrogate models, the FCN-8s \cite{fcn}, U-Net \cite{unet}, PSPNet \cite{pspnet}, and LinkNet \cite{linknet}, are adopted to evaluate the performance of different adversarial attack methods on semantic segmentation.

The detailed quantitative results are presented in Tables \ref{tab:vaihingen_trans} and \ref{tab:zurich_trans}. Compared to the scene classification models, the semantic segmentation models are generally more difficult to attack, as pixel-wise adversarial perturbations are required for the latter models. Nevertheless, it can be observed from Tables \ref{tab:vaihingen_trans} and \ref{tab:zurich_trans} that in all black-box attack scenarios, the proposed Mixup-Attack can still outperform the comparing attack methods by large margins. Take the results of FCN-8s$\rightarrow$SegNet on the Vaihingen dataset, for example. While C\&W obtains a success rate of about $38\%$, the proposed Mixup-Attack can achieve a success rate of over $71\%$, which is much higher with more than $30$ percentage points. Similar phenomena can be observed in the Zurich Summer dataset.

It can also be observed that while Mixcut-Attack can achieve competitive performance when attacking scene classification models, its success rate is relatively lower than that of Mixup-Attack for the semantic segmentation task. One possible reason is that mixcut samples only perturb the real spatial distribution of different pixels in the clean image but cannot change the pixel values. Thus, they could be helpful in fooling scene classification models where only one label is predicted for a whole image but would have less influence on semantic segmentation models, as they would make a pixel-wise classification for the whole image.

To evaluate how each module in the proposed methods influences the adversarial attack performance for the semantic segmentation task, the quantitative ablation study results are presented in Table \ref{tab:ablation_seg}. In both Vaihingen and Zurich Summer datasets, we can find that directly using the cross-entropy loss $\mathcal{L}_{ce}$ alone can hardly achieve a high success rate, while combining both $\mathcal{L}_{ce}$ and $\mathcal{L}_{mix}$ can significantly improve the performance by around $45$ and $34$ percentage points for Mixup-Attack, and $16$ and $26$ percentage points for Mixcut-Attack. Besides, the momentum strategy brings about relatively limited success rate improvement which is different from the scene classification scenario in Table \ref{tab:ablation_cls}.

\begin{table*}[]
\centering
\caption{Quantitative scene classification results of different deep neural networks on the clean and UAE-RS test sets.}
\label{tab:oa}
\resizebox{\textwidth}{!}{%
\begin{tabular*}{\linewidth}{@{\extracolsep{\fill}}ccccccc}
\hline
 & \multicolumn{3}{c}{UCM} & \multicolumn{3}{c}{AID} \\
Model & Clean Test Set & UAE-RS Test Set & OA Gap & Clean Test Set & UAE-RS Test Set & OA Gap \\ \hline
AlexNet \cite{alexnet} & 90.28 & 30.86 & -59.42 & 89.74 & 18.26 & -71.48 \\
VGG11 \cite{vgg} & 94.57 & 26.57 & -68.00 & 91.22 & 12.62 & -78.60 \\
VGG16 \cite{vgg} & 93.04 & 19.52 & -73.52 & 90.00 & 13.46 & -76.54 \\
VGG19 \cite{vgg} & 92.85 & 29.62 & -63.23 & 88.30 & 15.44 & -72.86 \\
Inception-v3 \cite{inception} & 96.28 & 24.86 & -71.42 & 92.98 & 23.48 & -69.50 \\
ResNet18 \cite{resnet} & 95.90 & 2.95 & -92.95 & 94.76 & 0.02 & -94.74 \\
ResNet50 \cite{resnet} & 96.76 & 25.52 & -71.24 & 92.68 & 6.20 & -86.48 \\
ResNet101 \cite{resnet} & 95.80 & 28.10 & -67.70 & 92.92 & 9.74 & -83.18 \\
ResNeXt50 \cite{resnext} & 97.33 & 26.76 & -70.57 & 93.50 & 11.78 & -81.72 \\
ResNeXt101 \cite{resnext} & 97.33 & 33.52 & -63.81 & 95.46 & 12.60 & -82.86 \\
DenseNet121 \cite{densenet} & 97.04 & 17.14 & -79.90 & 95.50 & 10.16 & -85.34 \\
DenseNet169 \cite{densenet} & 97.42 & 25.90 & -71.52 & 95.54 & 9.72 & -85.82 \\
DenseNet201 \cite{densenet} & 97.33 & 26.38 & -70.95 & 96.30 & 9.60 & -86.70 \\
RegNetX-400MF \cite{regnet} & 94.57 & 27.33 & -67.24 & 94.38 & 19.18 & -75.20 \\
RegNetX-8GF \cite{regnet} & 97.14 & 40.76 & -56.38 & 96.22 & 19.24 & -76.98 \\
RegNetX-16GF \cite{regnet} & 97.90 & 34.86 & -63.04 & 95.84 & 13.34 & -82.50 \\\hline
\end{tabular*}}
\\
\vspace{2pt}
\leftline{\scriptsize \quad Note: Results are reported in overall accuracy (\%).}
\end{table*}

\begin{table*}[]
\centering
\caption{Quantitative semantic segmentation results of different deep neural networks on the clean and UAE-RS test sets.}
\label{tab:mf1}
\resizebox{\textwidth}{!}{%
\begin{tabular*}{\linewidth}{@{\extracolsep{\fill}}ccccccc}
\hline
 & \multicolumn{3}{c}{Vaihingen} & \multicolumn{3}{c}{Zurich Summer} \\
Model & Clean Test Set & UAE-RS Test Set & Mean $F_1$ Gap & Clean Test Set & UAE-RS Test Set & Mean $F_1$ Gap \\ \hline
FCN-32s \cite{fcn}& 69.48 & 35.00 & -34.48 & 66.26 & 32.31 & -33.95 \\
FCN-16s \cite{fcn} & 69.70 & 27.02 & -42.68 & 66.34 & 34.80 & -31.54 \\
FCN-8s \cite{fcn} & 82.22 & 22.04 & -60.18 & 79.90 & 40.52 & -39.38 \\
DeepLab-v2 \cite{deeplabv2} & 77.04 & 34.12 & -42.92 & 74.38 & 45.48 & -28.90 \\
DeepLab-v3+ \cite{deeplabv3}& 84.36 & 14.56 & -69.80 & 82.51 & 62.55 & -19.96 \\
SegNet \cite{segnet}& 78.70 & 17.84 & -60.86 & 75.59 & 35.58 & -40.01 \\
ICNet \cite{icnet} & 80.89 & 41.00 & -39.89 & 78.87 & 59.77 & -19.10 \\
ContextNet \cite{contextnet} & 81.17 & 47.80 & -33.37 & 77.89 & 63.71 & -14.18 \\
SQNet \cite{sqnet} & 81.85 & 39.08 & -42.77 & 76.32 & 55.29 & -21.03 \\
PSPNet \cite{pspnet} & 83.11 & 21.43 & -61.68 & 77.55 & 65.39 & -12.16 \\
U-Net \cite{unet} & 83.61 & 16.09 & -67.52 & 80.78 & 56.58 & -24.20 \\
LinkNet \cite{linknet} & 82.30 & 24.36 & -57.94 & 79.98 & 48.67 & -31.31 \\
FRRNet-A \cite{frrnet} & 84.17 & 16.75 & -67.42 & 80.50 & 58.20 & -22.30 \\
FRRNet-B \cite{frrnet} & 84.27 & 28.03 & -56.24 & 79.27 & 67.31 & -11.96 \\ \hline
\end{tabular*}}
\\
\vspace{2pt}
\leftline{\scriptsize \quad Note: Results are reported in mean $F_1$ score (\%).}
\end{table*}

\subsection{UAE-RS Dataset}

Considering the high success rate of the proposed methods for the black-box adversarial attack, we collect the generated adversarial examples in a dataset named UAE-RS. To the best of our knowledge, UAE-RS is the first adversarial examples dataset in the remote sensing community, which may serve as a benchmark that helps researchers to design deep neural networks with strong resistance toward adversarial attacks.

To build UAE-RS, we use the Mixcut-Attack method to attack ResNet18 with 1050 test samples from the UCM dataset and 5000 test samples from the AID dataset for scene classification, and use the Mixup-Attack method to attack FCN8s with 5 test images from the Vaihingen dataset (image IDs: 11, 15, 28, 30, 34) and 5 test images from the Zurich Summer dataset (image IDs: 16, 17, 18, 19, 20)  for semantic segmentation.

\subsubsection{UAE-RS Samples on Scene Classification} Some example images in the original UCM and AID datasets and the corresponding adversarial images in the UAE-RS dataset are presented in Figs. \ref{fig:ucm} and \ref{fig:aid}. It can be observed that the generated adversarial images may look very similar to the original clean images to a human observer, despite their ability to seriously fool different deep neural networks into making wrong predictions.

We further evaluate the classification performance of different deep models on the original test images in UCM and AID datasets and the adversarial images in the UAE-RS dataset. The overall accuracy $OA=n_{correct}/n_{total}$ is utilized as the evaluation metric, where $n_{correct}$ denotes the number of correctly classified samples. The detailed quantitative results are presented in Table \ref{tab:oa}. We find that most of the models reported here can achieve an OA of more than $90\%$ in original clean test sets, which demonstrates the great capability of these deep neural networks. Nevertheless, the performance of all models drops significantly on the adversarial test set. The OA gap between the clean set and adversarial set can reach more than $70$ percentage points in many cases. These results indicate that adversarial examples in UAE-RS can easily mislead the existing state-of-the-art deep neural networks to make the wrong classification with high success rates on the scene classification of very high-resolution remote sensing images.

\subsubsection{UAE-RS Samples on Semantic Segmentation}
Some example images in the original Vaihingen and Zurich Summer datasets and the corresponding adversarial images in the UAE-RS dataset are presented in Figs. \ref{fig:vaihingen} and \ref{fig:zurich}. We also visualize the segmentation maps of SegNet on both the original clean images and the adversarial images. It can be observed that although there appears to be little difference between the generated adversarial images and the original clean images for the human eye, adversarial images can seriously fool SegNet into making wrong predictions. While the segmentation maps of SegNet are very close to the ground-truth annotations, most pixels in the adversarial images are misclassified as the low vegetation category in the Vaihingen dataset or the building category in the Zurich Summer dataset.

We further evaluate the segmentation performance of different deep models on the original test images in Vaihingen and Zurich Summer datasets and the adversarial images in the UAE-RS dataset. The mean $F_1$ score $mF_1=\frac{1}{n_v}\sum{F_1^{\left(k\right)}}_{k=1}^{n_v}$ is utilized as the evaluation metric, where $F_1^{\left(k\right)}$ denotes the $F_1$ score of the $k$th category and can be calculated as:
\begin{equation}
F_1^{\left(k\right)}=2\cdot\frac{{\rm precision}^{\left(k\right)}\cdot{\rm recall}^{\left(k\right)}}{{\rm precision}^{\left(k\right)}+{\rm recall}^{\left(k\right)}}.
\label{eq:f1}
\end{equation}

The detailed quantitative results are presented in Table \ref{tab:mf1}. We find that most of the models reported here can achieve a mean $F_1$ score of more than $70\%$ in original clean test sets. By contrast, the performance of all models drops significantly on the adversarial test set, especially for the Vaihingen dataset. The mean $F_1$ gap between the clean set and adversarial set can reach more than $30$ percentage points in many cases. These results indicate that adversarial examples in UAE-RS can also fool most of the existing state-of-the-art deep neural networks on the semantic segmentation of very high-resolution remote sensing images. Besides, we also find networks that make use of global context information generally possess stronger resistance towards adversarial attacks. For example, the ContextNet shows relatively smaller mean $F_1$ gaps in both Vaihingen and Zurich Summer datasets. This could be a possible direction for researchers developing a robust deep model to address the threat of adversarial attacks in the future.

\section{Conclusions and Discussions}
In this study, we analyze the universal adversarial examples in remote sensing data for the first time. Specifically, we propose the Mixup-Attack and Mixcut-Attack methods to conduct black-box adversarial attacks. While most of the existing methods directly attack the predicted logits, the proposed methods aim to attack the shallow layer of deep neural networks by minimizing the KL-divergence between features of the virtual image (i.e., mixup or mixcut samples) and the input image. Despite their simplicity, extensive experiments on four benchmark very high-resolution remote sensing image datasets demonstrate that the proposed methods can generate transferable adversarial examples that cheat most of the state-of-the-art deep neural networks in both scene classification and semantic segmentation tasks with high success rates.

The experimental results in this study also indicate that the networks with deeper architectures generally possess stronger resistance against the black-box adversarial attack. Besides, the networks that can make use of global context information show stronger resistance towards adversarial attacks for the semantic segmentation task (e.g., ContextNet). These two aspects could also be potential directions for researchers developing a robust deep model to address the threat of adversarial attacks in the future. Another intriguing phenomenon is that Mixcut-Attack can achieve higher success rates on the attacks of scene classification tasks, while Mixup-Attack is more threatening to the attacks of semantic segmentation tasks according to our experimental results. One possible reason is that mixcut samples only perturb the real spatial distribution of different pixels in the clean image but cannot change the pixel values. Thus, mixcut samples could be helpful in attacking the scene classification task where only one label is predicted for the whole image but would have less influence on the semantic segmentation task where pixel-wise classification is required.

We further collect the generated universal adversarial examples in the dataset named UAE-RS, which is the first dataset that provides black-box adversarial samples in the remote sensing field. We hope UAE-RS may serve as a benchmark for researchers developing adversarial defenses in the future. Despite the serious threat that adversarial examples have brought to deep learning models, it is also reported that training with adversarial examples could result in a better regularization ability and learn domain-invariant features \cite{na2020domain}. Thus, there also exists some potential for researchers to address the domain adaptation problem in remote sensing tasks via adversarial examples, which can further be investigated as a possible future work.

\section*{Acknowledgment}
The authors would like to thank Prof. Shawn Newsam for making the UCM dataset public available, Prof. Gui-Song Xia for providing the AID dataset, the International Society for Photogrammetry and Remote Sensing (ISPRS), and the German Society for Photogrammetry, Remote Sensing and Geoinformation (DGPF) for providing the Vaihingen dataset, and Dr. Michele Volpi for providing the Zurich Summer dataset.

\bibliographystyle{IEEEtran}

\bibliography{UAE}

\end{document}